\documentclass{article}

\usepackage[final]{nips_2016}

\usepackage[utf8]{inputenc} 
\usepackage[T1]{fontenc}     
\usepackage{hyperref}          
\usepackage{url}                  
\usepackage{booktabs}        
\usepackage{amsfonts}        
\usepackage{euscript}
\usepackage{amsmath}
\usepackage{mathrsfs}
\usepackage{amssymb}
\usepackage{nicefrac}          
\usepackage{microtype}       
\usepackage{graphicx}         
\usepackage[compact]{titlesec} 
\titlespacing{\section}{0pt}{0pt}{0pt}

\title{BaTFLED: Bayesian Tensor \\ Factorization Linked to External Data}

\author{
  Nathan H.~Lazar \\
  Oregon Health \& \\ Science University\\
  Portland, OR, USA \\
  \texttt{lazar@ohsu.edu} \\
  \And
  Mehmet G\"onen \\
  Ko\c c University \\
  Istanbul, Turkey \\
  \texttt{mehmetgonen@ku.edu.tr} \\
  \And
  Kemal S\"onmez \\
  Oregon Health \& \\ Science University\\
  Portland, OR, USA \\
  \texttt{sonmezk@ohsu.edu} \\
}

\begin{document}

\maketitle

\begin{abstract}
The vast majority of current machine learning algorithms are designed to predict single responses or a vector of responses, yet many types of response are more naturally organized as matrices or higher-order tensor objects where characteristics are shared across modes. We present a new machine learning algorithm BaTFLED (\textbf{Ba}yesian \textbf{T}ensor \textbf{F}actorization \textbf{L}inked to \textbf{E}xternal \textbf{D}ata) that predicts values in a three-dimensional response tensor using input features for each of the dimensions. BaTFLED uses a probabilistic Bayesian framework to learn projection matrices mapping input features for each mode into latent representations that multiply to form the response tensor. By utilizing a Tucker decomposition, the model can capture weights for interactions between latent factors for each mode in a small core tensor. Priors that encourage sparsity in the projection matrices and core tensor allow for feature selection and model regularization. This method is shown to far outperform elastic net and neural net models on 'cold start' tasks from data simulated in a three-mode structure. Additionally, we apply the model to predict dose-response curves in a panel of breast cancer cell lines treated with drug compounds that was used as a Dialogue for Reverse Engineering Assessments and Methods (DREAM) challenge. 
\end{abstract}

\section{The BaTFLED model}

\begin{figure}[h]
  \centering
    \includegraphics[width=.8\textwidth]{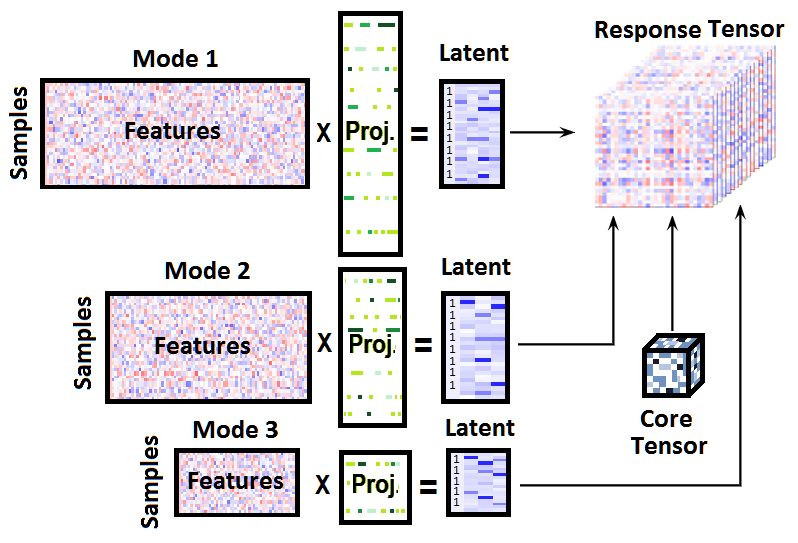}
  \caption{The BaTFLED model}
\end{figure}

\subsection{Model structure}

BaTFLED is a generative probabilistic model with the structure shown in figure 1. Matrices of input feature values for the training set examples are multiplied by learned projection matrices to form latent matrices. The number of columns in the latent matrices are set parameters determining the size of the latent space for that mode. The response tensor is factored using a Tucker decomposition (Tucker, 1964) into the latent matrices and a core tensor. This three-way matrix multiplication can be framed as a sum of outer products of the columns of the latent matrices weighted by values in the core tensor. In order to allow for lower-order interactions between latent factors and to obviate the need for normalization of responses, a column of ones is added to each latent matrix. Thus the core element corresponding to the outer product of these columns acts as a global constant added to all responses and products of other columns with the $1$s columns can encode interactions between only one or two of the modes. For comparison, we also implement a CANDECOMP/PARAFAC (CP) decomposition (Hitchcock, 1927), a special case of the Tucker decomposition where the latent space for each mode must be the same size and the core matrix only has values along the super-diagonal. 

In order to learn the values in the projection matrices, latent matrices and core tensor, we impose a probabilistic framework placing distributions on each of the unknown values. The distributions are chosen to maintain conjugacy and to encourage sparsity in the projection matrices and the core tensor. All of the major distributions are Gaussian and where sparsity is desired, means are set to zero and Gamma priors are placed on the precision (inverse variance). Setting the shape and scale for these Gamma distributions to extreme values (shape $\alpha=10^{-10}$, scale $\beta= 10^{10}$) encourages most of the variances to be near zero and the corresponding Gaussian to have a tight peak centered at zero. Variances are shared across rows of the projection matrices so that each feature is either selected or not for all the latent factors. 

\subsection{Training}

Due to size of the BaTFLED model, training would be computationally intractable with sampling methods, so instead we use a variational Bayesian approach. This seeks to maximize a lower bound of the posterior probability of the unknown values given the observed data and set parameters by approximating the full joint `p' distribution with a factorable `q' distribution. The parameters of each of these `q' distributions rely only on the expected values of other parameters in the model and equations for the optimal parameter values can be found analytically. Thus, once these update equations are derived, training proceeds by initializing all parameters randomly and updating each in sequence given the expectation of the other parameters. Full model details are in the supplemental material.

\subsection{Prediction}

Once trained, there are four types of prediction that this model can perform. First, if there are missing values in the response tensor, these can be filled in by multiplying the latent matrices through the core tensor. In this 'warm start' prediction we estimate a response for a combination when we have seen responses for other combinations involving the same input feature vectors. Second, given a vector of input features for a new example for one mode, 'cold start' predictions can be made for that example when combined with any example for the other two modes. Multiplying through the projection matrix and core tensor, a new matrix 'slice' of the response tensor is formed representing responses for this new test example across all training data. Similarly, 'cold start' prediction is possible for new exmples from two modes yielding a vector of responses across the third mode and lastly, 'cold start' prediction for all three modes produces a single predicted response value. 

\section{Experiments}

\subsection{Simulated data}

We first test the model on data simulated from the three-mode structure shown in figure 1 and compare it to baseline approaches. For these tests, response data is generated with 30 training examples and 100 features for each mode, 10 of which are used to produce the responses. Each mode has four latent factors, so with the added columns of ones, the core tensor is 5x5x5. Zero-centered Gaussian noise with a standard deviation of 1/10 of the response standard deviation is added to responses and each method is trained on 28 examples for each mode with 1\% of interactions removed for 'warm start' testing. 

In order to ascertain what types of interactions each of the tested models are able to discover, we generate data with three different sparse core tensors. The first core tensor only has non-zero values along the three primary edges, so the response tensor is formed using only 1D interactions between one latent factor from one mode and two columns of ones from the other modes. The second core tensor has non-zero values on three primary faces and so it allows 1D and 2D interactions between modes. The last core tensor can have non-zero values in any position, but is sparse so that only 1/2 of the possible interactions have non-zero weight. 

\subsection{Cancer cell line screening data}

Cancer cell line drug screen panels consist of a number of patient-derived cancer cell lines which are treated with a broad range of drugs at varying doses, grown for several days and imaged to assess cellular death. The resulting response data is inherently three-dimensional with one measurement for each combination of cell line, drug and dose. Current studies summarize these responses across doses by extracting parameters of fitted curves, and attempt to predict these values using genomic features. Instead, we predict response at each dose in the hope is that by incorporating more information, more useful relationships between the cell line genomics and drug structures can be revealed.

One such study was presented as a prediction challenge by the Dialogue on Reverse Engineering Assessment and Methods (DREAM) project (Costello et al., 2014). This challenge released data publicly so that teams from around the world could compete in the task of predicting responses in a panel of 52 breast cancer cell lines treated with 26 drugs. Although the original challenge was to predict a summary curve measure, data on response at each dose was also released and is utilized here. A set of 17 cell lines were held out as the final test set, but since the task of predicting responses for drugs was not part of the challenge, no drugs were separated into the test data set. In order to show performance on predicting multiple mode combinations, we present results both from 10 fold cross-validation runs on the 35 cell lines and 26 drugs in the training set (leaving out 4 cell lines and 3 drugs in each fold), as well as results for models trained with all of the original DREAM training data on the final 17 cell lines. The 733 input features for cell lines consist of binary indicators of cancer subtype, binary indicators of gene mutations and continuous measures of gene expression for a set of genes known to be associated with cancer (Bindal et al., 2011). The 433 input features for drugs consist of binary indicators of known gene targets, binary indicators of chemical substructures and continuous measures of 1D and 2D structural features extracted using the PaDEL software (Yap, 2011). 

\section{Results}

We examine the performance of the BaTFLED model both for predicting the responses and in selecting the true predictors and compare to four types of baseline models. In the first baseline, we predict the mean response across the training data averaging as many responses as possible for the the prediction task. All other models are run in ten replicates with input features consisting of vectors of the features for each mode concatenated together and a single continuous response for each combination of input examples. We train elastic net models using the R package `glmnet' (Friedman, et al. 2010) with different sparsity settings ranging from ridge regression ($\alpha=0$) to LASSO ($\alpha=1$). Random forest and neural net models are run in R using the `h2o' package (Aiello et al. 2016). Random forest models with 1,000 and 5,000 trees of depth 5, and 1,000 trees of depth 10 are tested as well as neural net models with 1, 2 and 3 hidden layers with 1,500, 750 and 500 nodes in each layer respectively. Neural net models use rectified linear activation functions with a dropout fraction of $0.2$ on the input layer and $0.5$ on hidden layers. Full results are given in the supplemental material.

On simulated data, BaTFLED models are run for $100$ iterations with prior parameters $\alpha=10^{-5}$ and $\beta=10^{5}$. The Tucker models are given $5$x$5$x$5$ cores and the CP models a latent dimension of $64$. All experiments were run in 10 replicates and we report the mean RMSE across replicates for representative models in table 1. Also shown are AUROC (area under the receiver operator characteristic) measures for these models when selecting predictors in data generated with 1,2 and 3D interactions. While BaTFLED performs comparably to baseline methods when predicting linear interactions it is able to learn higher-order relationships in data while other methods cannot. Additionally, BaTFLED selects the correct features in models with higher-order relationships at a higher frequency than other models.

For the DREAM data, we compare to the same baseline methods and report Pearson correlation measures in table 2. Since the DREAM data has more input features and less than half the total responses than the simulated data, we believe that it is somewhat underpowered for this prediction task. Simulation with artificially underpowered datasets indicate that this can be partially overcome by performing multiple rounds of testing to reduce the number of features. For the the cross-validation runs we found the best performance by first training BaTFLED models for 200 iterations with a $10$x$10$x$10$ core and strong sparsity priors ($\alpha=10^{-10}$ and $\beta=10^{10}$), keeping the union top 15\% of predictors across folds, and retraining for 40 iterations without encouraging sparsity. For the final DREAM testing runs, we tried a similar scheme, but found that a single round of training for 120 iterations with a $10$x$10$x$10$ core and strong sparsity priors performed as well as the second round. Our current results suggest that BaTFLED is able to predict responses as well or better than other baseline methods in both the cross-validation setting and on held-out test data on this very difficult challenge. In addition, BaTFLED may be able to discover interactions across modes that other methods cannot. 

\begin{table}[h]
  \caption{Mean normalized RMSE on a simulated data and AUROC for mode 1 over ten replicates .}
  \label{simulated}
  \centering
  \resizebox*{.9\textwidth}{!}{
  \begin{tabular}{lccccccccc}
    \toprule
         & Warm & \multicolumn{1}{p{.9cm}}{\centering Mode 1} 
         & \multicolumn{1}{p{.9cm}}{\centering Mode 2} 
         & \multicolumn{1}{p{.9cm}}{\centering Mode 3}
         & \multicolumn{1}{p{.9cm}}{\centering Mode \\ 1\&2} 
         & \multicolumn{1}{p{.9cm}}{\centering Mode \\ 1\&3} 
         & \multicolumn{1}{p{.9cm}}{\centering Mode \\ 2\&3} 
         & \multicolumn{1}{p{.9cm}}{\centering Mode \\ 1,2\&3} 
         & \multicolumn{1}{p{1.1cm}}{\centering Mode 1 \\ AUROC} \\
    \cmidrule{2-10} 
    & \multicolumn{9}{c}{Data generated with only 1D interactions} \\
    \cmidrule{2-10} 
    Mean            & 0.65 & 0.57 & 0.36 & 0.52 & 1.19 & 1.29 & 1.16 & 0.81 \\ 
    LASSO          & \textbf{0.10} & \textbf{0.14} & \textbf{0.18} & 0.27 & \textbf{0.18} & 0.31 & 0.30 & 0.30 & \textbf{0.93} \\ 
    Random forest  & 0.18 & 0.59 & 0.40 & 0.58 & 0.63 & 0.76 & 0.62 & 0.77 & 0.64 \\ 
    Neural net     & 0.65 & 0.77 & 0.68 & 0.72 & 0.73 & 0.85 & 0.82 & 0.84 & 0.72 \\ 
    BaTFLED CP.    & 0.38 & 0.46 & 0.42 & 0.46 & 0.45 & 0.52 & 0.45 & 0.46 & 0.87 \\ 
    BaTFLED Tucker       &  0.11 & \textbf{0.14} & 0.23 & \textbf{0.22} & 0.24 & \textbf{0.25} & \textbf{0.29} & \textbf{0.28} & 0.90 \\
    & \multicolumn{9}{c}{Data generated with 1D \& 2D interactions} \\
    \cmidrule{2-10} 
    Mean                 &  0.88 & 0.74 & 0.78 & 0.84 & 0.94 & 1.09 & 1.37 & 0.89 \\ 
    LASSO              & 0.95 & 0.86 & 0.93 & 1.05 & 0.81 & 0.86 & 1.13 & 0.88 & 0.74 \\ 
    Random forest  &0.42 & 0.69 & 0.73 & 0.93 & 0.75 & 0.82 & 1.05 & 0.90 & 0.73 \\ 
    Neural net         & 0.37 & 0.73 & 0.72 & 0.84 & 0.81 & 0.86 & 1.05 & 0.96 & 0.88 \\ 
    BaTFLED CP.      &  0.68 & 0.70 & 0.67 & 0.77 & 0.66 & 0.71 & 0.81 & 0.64 & 0.90 \\ 
    BaTFLED Tucker       & \textbf{0.11} & \textbf{0.12} & \textbf{0.11} & \textbf{0.11} & \textbf{0.12} & \textbf{0.12} & \textbf{0.12} & \textbf{0.12} & \textbf{0.98} \\ 
    & \multicolumn{9}{c}{Data generated with 1D, 2D \& 3D interactions} \\
    \cmidrule{2-10} 
    Mean                  & 0.98 & 1.02 & 0.80 & 1.05 & 0.85 & 1.13 & 0.89 & 0.81  \\ 
    LASSO                & 0.96 & 0.95 & 0.82 & 1.08 & 0.79 & 1.06 & 0.83 & 0.82 & 0.58 \\ 
    Random forest    & 0.61 & 0.92 & 0.78 & 1.05 & 0.81 & 1.06 & 0.84 & 0.85 & 0.75 \\ 
    Neural net           & 0.33 & 0.92 & 0.82 & 1.10 & 0.89 & 1.03 & 0.88 & 0.92 & 0.85 \\ 
    BaTFLED CP.       & 0.96 & 0.95 & 0.83 & 1.08 & 0.78 & 1.05 & 0.83 & 0.81 & 0.48 \\ 
    BaTFLED Tucker  & \textbf{0.12} & \textbf{0.12} & \textbf{0.12} & \textbf{0.12} & \textbf{0.12} & \textbf{0.13} & \textbf{0.12} & \textbf{0.13} &\textbf{1.00} \\ 
    \bottomrule
  \end{tabular}}
\end{table}

\begin{table}[h]
  \caption{Mean Pearson correlation on DREAM dataset.} 
  \label{DREAM}
  \centering
  \resizebox*{.9\textwidth}{!}{
  \begin{tabular}{lcccccccc}
    \toprule
            & Training & Warm & Cell lines & Drugs 
            & \multicolumn{1}{p{1.5cm}}{\centering Cl \& dr}
            & Training & Warm & Cell lines \\
    \cmidrule(lr){2-6} 
    \cmidrule(lr){7-9}
    & \multicolumn{5}{c}{Cross-validation on training dataset} & 
    \multicolumn{3}{c}{Final test set} \\
    \cmidrule(lr){2-6} 
    \cmidrule(lr){7-9}
    Mean                 & 0.65 & 0.40 & 0.63 & 0.66 & 0.43 & 0.65 & 0.46 & 0.57 \\ 
    LASSO               & 0.79 & 0.77 & 0.62 & 0.66 & 0.48 & 0.79 & 0.75 & 0.59 \\ 
    Random forest   & 0.87 & \textbf{0.82} & 0.58 & 0.66 & 0.42 & 0.87 & \textbf{0.78} & 0.58   \\ 
    Neural net          & 0.94 & 0.78 & 0.53 & 0.58 & 0.39 & 0.93 & 0.77 & 0.60 \\ 
    BaTFLED Tucker* & \textbf{0.96} & 0.75 & \textbf{0.70} & \textbf{0.71} & \textbf{0.57} & \textbf{0.96} & 0.76 & \textbf{0.61} \\ 
    \bottomrule
    \multicolumn{9}{l}{* Cross-validation results are from a two-round training strategy.} \\
  \end{tabular}}
\end{table}

\section{Conclusions}

As technological advancements in biology allow for the characterization of systems across an increasing number of molecular, structural, and behavioral dimensions, methods that can combine such datasets in a principled and interpretable manor are greatly needed. Tensor factorizations provide a natural framework for organizing such data and this work demonstrates the usefulness of these methods in a predictive setting. BaTFLED is designed for general use, is freely available and can be easily applied to many different applications. A fully functional R package is available at \url{https://www.github.com/nathanlazar/BaTFLED3D}.

\newpage

\subsubsection*{Acknowledgments}

Thanks to Dr. Shannon McWeeney, Dr. Adam Margolin and Dr. Lucia Carbone.

\section*{References}

Aiello, S., Kraljevic, T., Maj, P. and with contributions from the H2O.ai team (2016). h2o: R Interface for H2O. R package version 3.10.0.8. https://CRAN.R-project.org/package=h2o

Bindal, N., Forbes, S.A., Beare, D., Gunasekaran, P., Leung, K., Kok, C.Y., Jia, M., Bamford, S., Cole, C., Ward, S., et al. (2011). COSMIC: the catalogue of somatic mutations in cancer. {\it Genome Biology} {\bf 12}: P3.

Costello, J.C., Heiser, L.M., Georgii, E., Gönen, M., Menden, M.P., Wang, N.J., Bansal, M., Ammad-ud-din, M., Hintsanen, P., Khan, S.A., et al. (2014). A community effort to assess and improve drug sensitivity prediction algorithms. {\it Nature Biotechnology} {\bf 32}: 1202–1212.

Friedman, J., Hastie, T.\ \& Tibshirani R.\ (2010). Regularization Paths for Generalized Linear Models via Coordinate Descent. {\it Journal of Statistical Software} {\bf 33}(1): 1-22. URL http://www.jstatsoft.org/v33/i01/

Hitchcock, F.L. (1927). {\it The expression of a tensor or a polyadic as a sum of products} (Cambridge, Mass.: Inst. of Technology).

Tucker, L.R. (1964). The extension of factor analysis to three-dimensional matrices. In  H.\ Gulliksen, and N.\ Frederiksen (eds.), {\it Contributions to Mathematical Psychology.}, pp.\ 110--127 New York: Holt, Rinehart and Winston.

Yap, C.W. (2011). PaDEL-descriptor: An open source software to calculate molecular descriptors and fingerprints. {\it Journal of Computational Chemistry} {\bf 32}: 1466–1474.

\newpage

\title{BaTFLED: Bayesian Tensor Factorization Linked to External Data (supplementary material)}

\maketitle

\section{Details of the BaTFLED model}

\subsection{Notation} 
The notation throughout this paper mostly follows the conventions of the excellent review (Kolda and Bader, 2007). I will use boldface capital letters to refer to tensors, capital letters to refer to matrices (and index upper limits), boldface lower case letters to refer to vectors and lower-case letters to refer to scalars. Indices will typically range from $1$ to their capital version so, for example, the three modes of the third-order response tensor $\textbf{Y}$ have indices $i$,$j$ and $k$ where $i=1…I$, $j=1…J$ and $k=1…K$ and a typical element will be written $y_{ijk}$. I use $X$ for the input feature data, $A$ for the projection matrices, $\Lambda$ for the matrix of priors on the precision (inverse variance) of the projection matrices, $H$ for the latent matrices, and $\textbf{C}$ for the core tensor. The superscripts indicate the three modes ($c$, $d$ and $s$ to suggest cell lines, drugs and doses) and core elements ($r$). For simplicity we denote all the observed values and set parameters (except the responses) by $\Xi$.
\begin{align} 
\Xi &= \{ X^c, X^d, X^s, \sigma^2, \sigma_c^2, \sigma_d^2, \sigma_s^d, \alpha^c, \alpha^d, \alpha^s, \alpha^r, \beta^c, \beta^d, \beta^s, \beta^r, \delta_{ijk} \}
\end{align}
and all random variables by $\Theta$
\begin{align}
\Theta = \{ H^c, H^d, H^s, A^c, A^d, A^s, \Lambda^c, \Lambda^d, \Lambda^s, \boldsymbol{\Lambda}^r\}
\end{align}

\subsection{Probabilistic model}

The full joint distribution of the the prior parameters $\Lambda^c, \Lambda^d, \Lambda^s, \boldsymbol{\Lambda}^r$, projection matrices $A^c, A^d, A^s$, latent factor matrices $H^c, H^d, H^s$ and the core tensor $\textbf{C}$ given the observed data $\textbf{Y}, X^c, X^d, X^s, \delta_{ijk}$ and the set parameters $\sigma^2, \sigma_c^2, \sigma_d^2, \sigma_s^2, \alpha^c, \alpha^d, \alpha^s, \alpha^r$ $ \beta^c, \beta^d, \beta^s, \beta^r$ is summarized as $p(\Theta, \textbf{Y} | \Xi)$.
This distribution factors as follows (dependencies on set parameters are omitted for clarity).
\begin{align}
p(\Theta,\textbf{Y} | \Xi)&= p(\textbf{Y}, H^c, H^d, H^s, A^c, A^d, A^s, \Lambda^c, \Lambda^d, \Lambda^s, \textbf{C}, \boldsymbol{\Lambda}^r | X^c, X^d, X^s, \delta_{ijk}) \nonumber\\
&= p(\textbf{Y}|\textbf{C},H^c,H^d,H^s, \delta_{ijk}) p(H^c|A^c,X^c) p(H^d|A^d,X^d) p(H^s|A^s,X^s) \nonumber \\
 &\qquad p(\textbf{C}|\boldsymbol{\Lambda^r}) p(A^c|\Lambda^c)p(A^d|\Lambda^d)p(A^s|\Lambda^s)p(\Lambda^c)p(\Lambda^d)p(\Lambda^s) \label{fulljoint}
\end{align}
The distributional assumptions for each of the factors are

\begin{align}
p(\textbf{Y}|\textbf{C},H^c,H^d,H^s) &= \prod_{i=1}^I \prod_{j=1}^J \prod_{k=1}^K \mathcal{N}(y_{ijk}; 
  \sum_{r_1=1}^{R_1} \sum_{r_2=1}^{R_2} \sum_{r_3=1}^{R_3} c_{r_1r_2r_3} h_{ir_1}^c h_{jr_2}^d h_{kr_3}^s,\sigma^2)^{\delta_{ijk}} \label{pofy} \\
p(H^{c}|A^{c},X^{c}) &= \prod_{i=1}^I \prod_{r_1=1}^{R_1} \mathcal{N} (h_{ir_1}^c ; \textbf{x}_i^c \textbf{a}_{r_1}^c , \sigma_c^2) \label{pofHc} \\ 
p(H^{d}|A^{d},X^{d}) &= \prod_{j=1}^J \prod_{r_2=1}^{R_2} \mathcal{N} (h_{jr_2}^d ; \textbf{x}_j^d \textbf{a}_{r_2}^d , \sigma_d^2) \label{pofHd} \\
p(H^{s}|A^{s},X^{s}) &= \prod_{k=1}^K \prod_{r_3=1}^{R_3} \mathcal{N} (h_{kr_3}^s ; \textbf{x}_k^s \textbf{a}_{r_3}^s , \sigma_s^2) \label{pofHs} \\
p(\textbf{C}|\boldsymbol{\Lambda}^r) &= \prod_{r_1=1}^{R_1} \prod_{r_2=1}^{R_2} \prod_{r_3=1}^{R_3} 
  \mathcal{N} (c_{r_1r_2r_3} ; 0 , (\lambda_{r_1r_2r_3}^{r})^{-1}) \label{pofC} \\
p(A^c|\Lambda^c ) &= \prod_{p=1}^P \prod_{r_1=1}^{R_1} \mathcal{N}(a_{pr_1}^{c} ; 0, (\lambda_{pr_1}^{c})^{-1}) \label{pofAc} \\
p(A^d|\Lambda^d ) &= \prod_{q=1}^Q \prod_{r_2=1}^{R_2} \mathcal{N}(a_{qr_2}^{d} ; 0, (\lambda_{qr_2}^{d})^{-1}) \label{pofAd} \\
p(A^s|\Lambda^s ) &= \prod_{t=1}^T \prod_{r_3=1}^{R_3} \mathcal{N}(a_{tr_3}^{s} ; 0, (\lambda_{tr_3}^{s})^{-1}) \label{pofAs} \\
p(\boldsymbol{\Lambda}^r) &= \prod_{r_1=1}^{R_1} \prod_{r_2=1}^{R_2} \prod_{r_3=1}^{R_3} \mathcal{G}(\lambda_{r_1r_2r_3}^r ; \alpha^r, \beta^r) \label{poflambdar} \\
p(\Lambda^c) &= \prod_{p=1}^P \prod_{r_1=1}^{R_1} \mathcal{G}(\lambda_{pr_1}^c ; \alpha^c, \beta^c) \label{poflambdac} \\
p(\Lambda^d) &= \prod_{q=1}^Q \prod_{r_2=1}^{R_2} \mathcal{G}(\lambda_{qr_2}^d ; \alpha^d, \beta^d) \label{poflambdad}\\
p(\Lambda^s) &= \prod_{t=1}^T \prod_{r_3=1}^{R_3} \mathcal{G}(\lambda_{tr_3}^s ; \alpha^s, \beta^s) \label{poflambdas}\\
\delta_{ijk} &= \begin{cases} 1 & if~y_{ijk}~is~observed  \\
  0 & otherwise \end{cases} \nonumber \\
\end{align}
Where $\mathcal{N}(x;\mu,\sigma^2)$ is a normal distribution with mean $\mu$ and variance $\sigma^2$ and $\mathcal{G}(x;a,b)$ is a gamma distribution with shape $a$ and scale $b$.
We use a shared variance $\sigma^2$ for the response tensor and shared variances $\sigma^2_c$, $\sigma^2_d$ and $\sigma^2_s$ for the elements of the latent matrices $H^c, H^d$ and $H^s$ respectively. The gamma distributions on the $\lambda$ parameters for the projection matrices $A^c, A^d$ and $A^s$ and the core $\textbf{C}$ allows the user to encourage sparsity in these parts of the model. By setting the prior shape and scale parameters to extreme values (ex. $\alpha=10^{-10}$ and $\beta=10^{10}$) the majority of $\lambda$ values move toward zero as the model is trained. The precision values in the projection ($A$) matrices can optionally be shared across rows which encourages the use of the same predictor variables for all latent factors. 

Also, columns of ones can be introduced into the input ($X$) matrices and latent ($H$) matrices. In the input matrices this allows for bias terms, and in the a  latent matrices, these columns allow the model to learn marginal coefficients that apply to lower-dimensional subsets of the response tensor. For example if all three $H$ matrices have a constant, we extend the ranges of $r_1$, $r_2$ and $r_3$ to include a zero index and the corresponding element of the core matrix $(c_{000})$ can learn the 'intercept' for the response tensor; a value that is added to all responses. Similarly the $c_{100}$ element is a coefficient for the first latent factor for the first mode and encodes the one-dimensional effects that this latent factor will have on response regardless of the other two modes. The coefficient $c_{110}$ encodes interactions between the first latent factors of modes one and two that occur regardless of mode three.

\section{Inference}

With the model established above, our goal is to find the posterior distribution of variables in the model given the observations $p(\Theta | \Xi, \textbf{Y})$ as well as the marginal likelihood or model evidence $p(\textbf{Y} | \Xi)$. Exact inference of the posterior is not analytically solvable and sampling based approaches like Markov chain monte carlo (MCMC) would be computationally prohibitive for a model of this size. Instead we use a variational approximation (or variational Bayes) approach that maximizes a lower bound on the log of the posterior. This lower bound can be written as: 

\begin{equation}
\mathcal{L} = \mathbb{E}_{q(\Theta)}[\log p(\textbf{Y},\Theta | \Xi)] - \mathbb{E}_{q(\Theta)}[\log q(\Theta)].
\end{equation}

The new $q$ distribution is a joint distribution of the same variables as the $p$ distribution that approximates the $p$ distribution under the assumption that $q$ is factorable in some convenient way. Although the $q$ distribution has the same unknown variables as $p$, the parameters of these distributions (mean, variance, etc.) are different than those given above. Closed forms for these parameters can be derived analytically and they depend on expectations over the $q$ distributions of other variables in the model. Thus, in order to find the $q$ distribution that best approximates the $p$ distribution an iterative expectation maximization process is employed. 

The joint likelihood $p(\textbf{Y},\Theta | \Xi)$ is given in (\ref{fulljoint}) and we assume that the $q$ distribution is fully factorable:

\begin{align}
q(\Theta) = q(H^c)q(H^d)q(H^s)q(A^c)q(A^d)q(A^s)q(\boldsymbol{\lambda}^c)q(\boldsymbol{\lambda}^d)q(\lambda^s)q(\boldsymbol{\tau}^c)q(\boldsymbol{\tau}^d)q(\boldsymbol{\tau}^s)
\end{align}

Thus the lower bound can be written:
\begin{align}
\mathcal{L} &= \mathbb{E}_{q(H^c)q(H^d)q(H^s)}[\log p(\textbf{Y}|H^c,H^d,H^s)] \nonumber \\
 &+ \mathbb{E}_{q(H^c)q(A^c)}[p(H^c|A^c,X^c)] \nonumber \\
 &+  \mathbb{E}_{q(H^d)q(A^d)}[\log p(H^d|A^d,X^d)] \nonumber \\
 &+  \mathbb{E}_{q(H^s)q(A^s)} [\log p(H^s|A^s,X^s)] \nonumber \\
 &+ \mathbb{E}_{q(A^c)q(\boldsymbol{\lambda}^c)} [ \log p(A^c|\boldsymbol{\lambda}^c)]
 +  \mathbb{E}_{q(A^d)q(\boldsymbol{\lambda}^d)} [ \log p(A^d|\boldsymbol{\lambda}^d)]
 +  \mathbb{E}_{q(A^s)q(\boldsymbol{\lambda}^s)} [ \log p(A^s|\boldsymbol{\lambda}^s)] \nonumber \\
 &+ \mathbb{E}_{q(\boldsymbol{\lambda}^c)q(\boldsymbol{\tau}^c)} [\log p(\boldsymbol{\lambda}^c|\boldsymbol{\tau}^c)]
 +  \mathbb{E}_{q(\boldsymbol{\lambda}^d)q(\boldsymbol{\tau}^d)} [\log p(\boldsymbol{\lambda}^d|\boldsymbol{\tau}^d)]
 +  \mathbb{E}_{q(\boldsymbol{\lambda}^s)q(\boldsymbol{\tau}^s)} [\log p(\boldsymbol{\lambda}^s|\boldsymbol{\tau}^s)] \nonumber \\
 &+ \mathbb{E}_{q(\boldsymbol{\tau}^c)} [\log p(\boldsymbol{\tau}^c)]
 +  \mathbb{E}_{q(\boldsymbol{\tau}^d)} [\log p(\boldsymbol{\tau}^d)]
 +  \mathbb{E}_{q(\boldsymbol{\tau}^s)} [\log p(\boldsymbol{\tau}^s)] \nonumber \\
 &- \mathbb{E}_{q(H^c)}[\log q(H^c)] 
 -  \mathbb{E}_{q(H^d)}[\log q(H^d)] 
 -  \mathbb{E}_{q(H^s)}[\log q(H^s)] \nonumber \\ 
 &- \mathbb{E}_{q(A^c)}[\log q(A^c)]
 -  \mathbb{E}_{q(A^d)}[\log q(A^d)]
 -  \mathbb{E}_{q(A^s)}[\log q(A^s)] \nonumber \\
 &- \mathbb{E}_{q(\boldsymbol{\lambda}^c)} [\log q(\boldsymbol{\lambda}^c)] 
 -  \mathbb{E}_{q(\boldsymbol{\lambda}^d)} [\log q(\boldsymbol{\lambda}^d)] 
 -  \mathbb{E}_{q(\boldsymbol{\lambda}^s)} [\log q(\boldsymbol{\lambda}^s)] \nonumber \\
 &- \mathbb{E}_{q(\boldsymbol{\tau}^c)} [\log q(\boldsymbol{\tau}^c)] 
 -  \mathbb{E}_{q(\boldsymbol{\tau}^d)} [\log q(\boldsymbol{\tau}^d)] 
 -  \mathbb{E}_{q(\boldsymbol{\tau}^s)} [\log q(\boldsymbol{\tau}^s)]
\end{align}

The parameters for the $q$ distributions of a given variable $\theta$  that maximize this lower bound can be obtained by finding the expectation of $\log p(\Theta|\Xi)$ with respect to the q distributions over all other variables (denoted $\Theta \backslash \theta$). This gives expressions (update equations) for each of the $q$ distributions that depend only on moments of the other variables (and fixed parameters). 

\begin{align}
\log q(\theta) &= \mathbb{E}_{q(\Theta \backslash \theta)} \left[ \log p(\Theta | \Xi) \right]
\end{align}

\subsection{Update equations} \label{update_equations}

The optimal $q$ distributions for individual elements of the first mode and core elements are given below, other modes are identical. All expectations are over the $q$ distributions.

Prior $\Lambda$ matrices:

\begin{equation}
  q(\lambda_{pr_1}^c) = \mathcal{G} \left(\lambda_{pr_1}^c; \alpha^c + \frac{1}{2} , 
    \left( \frac{1}{2} \mathbb{E}[a_{pr_1}^c]^2 + \frac{1}{2}var(a_{pr_1}^c) + \frac{1}{\beta^c} \right)^{-1} \right) 
\end{equation}

\begin{equation}
q(\lambda_{r_1r_2r_3}^r) = \mathcal{G} \left(\lambda_{r_1r_2r_3}^r; \alpha^r + \frac{1}{2} , 
  \left( \frac{1}{2} \mathbb{E}[c_{r_1r_2r_3}]^2 + \frac{1}{2}var(c_{r_1r_2r_3}) + \frac{1}{\beta^r} \right)^{-1} \right) 
\end{equation}

Projection $A$ matrices

\begin{equation}
  q(\textbf{a}_{r_1}^c) = \mathcal{N} \left( \textbf{a}_{r_1}^c ; \boldsymbol{\mu}(\textbf{a}_{r_1}^c) , \Sigma (\textbf{a}_{r_1}^c) \right) \\
\end{equation}

with

\begin{align*}
  \mu (\textbf{a}_{r_1}^c) &= \left(\frac{1}{\sigma_c^2} \mathbb{E} [\textbf{h}_{r_1}^c]^{T} X^c \right) \Sigma (\textbf{a}_{r_1}^c) \\
  \Sigma (\textbf{a}_{r_1}^c) &= \left( \frac{1}{\sigma_c^2} (X^c)^T X^c + \mathbb{E} [ \boldsymbol{\lambda}_{r_1}^c] I \right)^{-1}
\end{align*}

Latent $H$ matrices:

\begin{equation}
q(h_{ir}^c)  = \mathcal{N}(h_{ir}^c;\mu (h_{ir}^c), \Sigma (h_{ir}^c)) \\
\end{equation}

with

\begin{align*}
\mu (h_{ir_1}^c) &= \Sigma (h_{ir_1}^c) \left( \frac{1}{\sigma^2} \sum_{j=1}^J \sum_{k=1}^K \delta_{ijk} \left[
  y_{ijk} \sum_{r_2=1}^{R_2} \sum_{r_3=1}^{R_3} \mathbb{E} [c_{r_1r_2r_3}] \mathbb{E} [h^d_{jr_2}] \mathbb{E} [h^s_{kr_3}] \right. \right. \\
  & - \sum_{r^*_1 \neq r_1} \sum_{r_2}^{R_2} \sum_{r_3}^{R_3} \mathbb{E} [c_{r_1r_2r_3}] \mathbb{E}[h^c_{ir^*_1}] \left[ 
    \sum_{r^*_2 \neq r_2} \sum_{r^*_3 \neq r_3} \mathbb{E}[c_{r^*_1r^*_2r^*_3}] \mathbb{E}[h^d_{jr_2}] \mathbb{E}[h^d_{jr^*_2}]
    \mathbb{E}[h^s_{kr_3}] \mathbb{E}[h^s_{kr^*_3}]  \right. \\
  & \hspace{1in} + \sum_{r^*_2 \neq r_2} \mathbb{E}[c_{r^*_1r^*_2r_3}] \mathbb{E}[h^d_{jr_2}] \mathbb{E}[h^d_{jr^*_2}]
    \left( \mathbb{E}[h^s_{kr_3}]^2 + Var(h^s_{kr_3}) \right) \\
  & \hspace{1in} + \sum_{r^*_3 \neq r_3} \mathbb{E}[c_{r^*_1r_2r^*_3}] \left( \mathbb{E}[h^d_{jr_2}]^2 + Var(h^d_{jr_2}) \right)
  \mathbb{E}[h^s_{kr_3}] \mathbb{E}[h^s_{kr^*_3}] \\
  & \hspace{1in} + \mathbb{E}[c_{r^*_1r_2r_3}] \left( \mathbb{E}[h^d_{jr_2}]^2 + Var(h^d_{jr_2}) \right) 
    \left( \mathbb{E}[h^s_{kr_3}]^2 + Var(h^s_{kr_3}) \right) \Bigg] \Bigg] \\
  & + \frac{1}{\sigma_c^2} \textbf{x}^c_i \textbf{a}^c_{r_1} \Bigg)
\end{align*}

and

\begin{align*}
\Sigma (h_{ir}^c) = & \Bigg( \frac{1}{\sigma^2} \sum_{j=1}^J \sum_{k=1}^{K} \delta_{ijk} \sum_{r_2=1}^{R_2} \sum_{r_3}^{R_3} \Bigg[
  \sum_{r^*_2 \neq r_2} \sum_{r^*_3 \neq r_3} \mathbb{E} [c_{r_1r_2r_3}] \mathbb{E} [c_{r_1r^*_2r^*_3}] 
    \mathbb{E} [h^d_{jr_2}] \mathbb{E} [h^d_{jr^*_2}] \mathbb{E} [h^s_{kr_3}] \mathbb{E} [h^s_{kr^*_3}] \\
  & \hspace{.1in} + \sum_{r^*_2 \neq r_2} \mathbb{E} [c_{r_1r_2r_3}] \mathbb{E} [c_{r_1r^*_2r_3}] 
    \mathbb{E} [h^d_{jr_2}] \mathbb{E} [h^d_{jr^*_2}] \left( \mathbb{E} [h^s_{kr_3}]^2 + Var(h^s_{kr_3}) \right) \\
  & \hspace{.1in} + \sum_{r^*_3 \neq r_3} \mathbb{E} [c_{r_1r_2r_3}] \mathbb{E} [c_{r_1r_2r_3^*}] 
    \left( \mathbb{E} [h^d_{jr_2}]^2 + Var(h^d_{jr_2}) \right) \mathbb{E} [h^s_{kr_3}] \mathbb{E} [h^s_{kr^*_3}]  \\
  & \hspace{.1in} + \left( \mathbb{E} [c_{r_1r_2r_3}]^2 + Var(c_{r_1r_2r_3}) \right) 
    \left( \mathbb{E} [h^d_{jr_2}]^2 + Var(h^d_{jr_2}) \right) \left( \mathbb{E} [h^s_{kr_3}]^2 + Var(h^s_{kr_3}) \right) \Bigg]
  + \frac{1}{\sigma_c^2} \Bigg)^{-1} 
\end{align*}

Core $\textbf{C}$ tensor

\begin{equation}
q(c_{r_1r_2r_3}) = \mathcal{N} \left(c_{r_1r_2r_3}; \mu(c_{r_1r_2r_3}), \Sigma(c_{r_1r_2r_3}) \right) \\
\end{equation}

with 

\begin{align*}
\mu(c_{r_1r_2r_3}) =& \Sigma(c_{r_1r_2r_3})  \frac{1}{\sigma^2} \sum_{i=1}^I \sum_{j=1}^J \sum_{k=1}^K \Biggl[
  \delta_{ijk} y_{ijk} \mathbb{E}[h^c_{ir_1}] \mathbb{E}[h^d_{jr_2}] \mathbb{E}[h^s_{kr_3}]  \\
  &+ \mathbb{E} [h^c_{ir_1}] \mathbb{E} [h^d_{jr_2}] \mathbb{E} [h^s_{kr_3}] 
  \sum_{r^*_1 \neq r_1} \sum_{r^*_2 \neq r_2} \sum_{r^*_3 \neq r_3} \mathbb{E}[c_{r^*_1r^*_2r^*_3}] 
  \mathbb{E} [h^c_{ir^*_1}] \mathbb{E} [h^d_{jr^*_2}] \mathbb{E} [h^s_{kr^*_3}] \\
  &+ \left( \mathbb{E} [h^c_{ir_1}]^2 + Var(h^c_{ir_1}) \right) \mathbb{E} [h^d_{jr_2}] \mathbb{E} [h^s_{kr_3}] 
  \sum_{r^*_2 \neq r_2} \sum_{r^*_3 \neq r_3} \mathbb{E}[c_{r_1r^*_2r^*_3}] \mathbb{E} [h^d_{jr^*_2}] \mathbb{E} [h^s_{kr^*_3}] \\
  &+ \mathbb{E} [h^c_{ir_1}] \left(\mathbb{E} [h^d_{jr_2}]^2 + Var(h^d_{jr_2}) \right) \mathbb{E} [h^s_{kr_3}] 
  \sum_{r^*_1 \neq r_1} \sum_{r^*_3 \neq r_3} \mathbb{E}[c_{r^*_1r_2r^*_3}] \mathbb{E} [h^c_{ir^*_1}] \mathbb{E} [h^s_{kr^*_3}] \\
  &+ \mathbb{E} [h^c_{ir_1}] \mathbb{E} [h^d_{jr_2}] \left( \mathbb{E} [h^s_{kr_3}]^2 + Var(h^s_{kr_3}) \right) 
  \sum_{r^*_1 \neq r_1} \sum_{r^*_2 \neq r_2} \mathbb{E}[c_{r^*_1r^*_2r_3}] \mathbb{E} [h^c_{ir^*_1}] \mathbb{E} [h^d_{jr^*_2}] \\
  &+ \left( \mathbb{E} [h^c_{ir_1}]^2 + Var(h^c_{ir_1}) \right) \left(\mathbb{E} [h^d_{jr_2}]^2 + Var(h^d_{jr_2}) \right) \mathbb{E} [h^s_{kr_3}] 
  \sum_{r^*_3 \neq r_3} \mathbb{E}[c_{r_1r_2r^*_3}] \mathbb{E} [h^s_{kr^*_3}] \\
  &+ \left( \mathbb{E} [h^c_{ir_1}]^2 + Var(h^c_{ir_1}) \right) \mathbb{E} [h^d_{jr_2}] \left( \mathbb{E} [h^s_{kr_3}]^2 + Var(h^s_{kr_3}) \right)
  \sum_{r^*_2 \neq r_2} \mathbb{E}[c_{r_1r^*_2r_3}] \mathbb{E} [h^d_{jr^*_2}] \\
  &+ \mathbb{E} [h^c_{ir_1}] \left(\mathbb{E} [h^d_{jr_2}]^2 + Var(h^d_{jr_2}) \right) \left( \mathbb{E} [h^s_{kr_3}]^2 + Var(h^s_{kr_3}) \right)
  \sum_{r^*_1 \neq r_1} \mathbb{E}[c_{r^*_1r_2r_3}] \mathbb{E} [h^c_{ir^*_1}] \Biggr] \\
\end{align*}

\begin{align*}
\Sigma(c_{r_1r_2r_3}) &= \Bigg( \frac{1}{\sigma^2} \sum_{i=1}^I \sum_{j=1}^J \sum_{k=1}^K \delta_{ijk} \left(\mathbb{E} [h^c_{ir_1}]^2 + Var(h^c_{ir_1}) \right) 
  \left(\mathbb{E} [h^d_{jr_2}]^2 + Var(h^d_{jr_2}) \right) \\
  & \hspace{1.4in} \left( \mathbb{E} [h^s_{kr_3}]^2 + Var(h^s_{kr_3}) \right) + \mathbb{E} [\lambda^r_{r_1r_2r_3}] \Bigg)^{-1} \\
\end{align*}

\newpage
\section{Experimental results}

Results for additional experiments and performance measures are given below.

\subsection{Simulated data}
Responses are simulated from the structure assumed by the BaTFLED model. Each of the three modes has 30 samples, 4 latent factors and 100 features, 10 of which influence response. Two samples for each mode are held out as validation data. For the `1D' interaction responses, the core only has values along the edges corresponding to the constant columns in the latent matrices. This means that no interactions between modes influence responses. For the `1D \& 2D' interaction responses, the core only has values in the matrix slices corresponding to the constant columns, so responses are built from interactions between at most two of the modes. For the `1D ,2D \& 3D' interaction responses, the core has values throughout, but is sparse so half of potential interactions have weight zero and do not contribute to the responses.

BaTFLED Tucker models are trained for 100 iterations with a $5$x$5$x$5$ core and sparsity priors $\alpha=10^{-5}$ and $\beta=10^5$ on both the projection matrices and the core. BaTFLED CP models are trained for 100 iterations with $64$ latent features and sparsity priors $\alpha=10^{-5}$ and $\beta=10^5$ on the projection matrices. The features for elastic net, neural net and random forest models are vectors consisting of the features for each mode concatenated. Elastic net models are trained using the R package `glmnet' (Friedman et al. 2010) and we choose lambda values that give the most regularized model within one standard deviation of the model with the lowest cross-validated error. Neural net models are trained using the `h2o' R package (Aiello et al. 2016) with $1$, $2$ and $3$ hidden layers with $1500$, $750$ and $500$ nodes in each layer respectively. Activations are rectified linear functions and a dropout fraction of $0.2$ on the input layer and $0.5$ on the hidden layers are used. For each replication, convergence is determined by monitoring the mean squared error (MSE) on ten cross-validation folds and stopping when the MSE changes by less than $0.01$ across five epochs. Random forest models are also trained in R using the h2o package. Splitting is determined using the the automatic option `mtries=-1' and models with $1,000$ trees of depth $5$, $5000$ trees of depth $5$, and $1000$ trees of depth $10$ are tested.

\begin{figure}[!htbp]
  \centering
    \includegraphics[width=\textwidth]{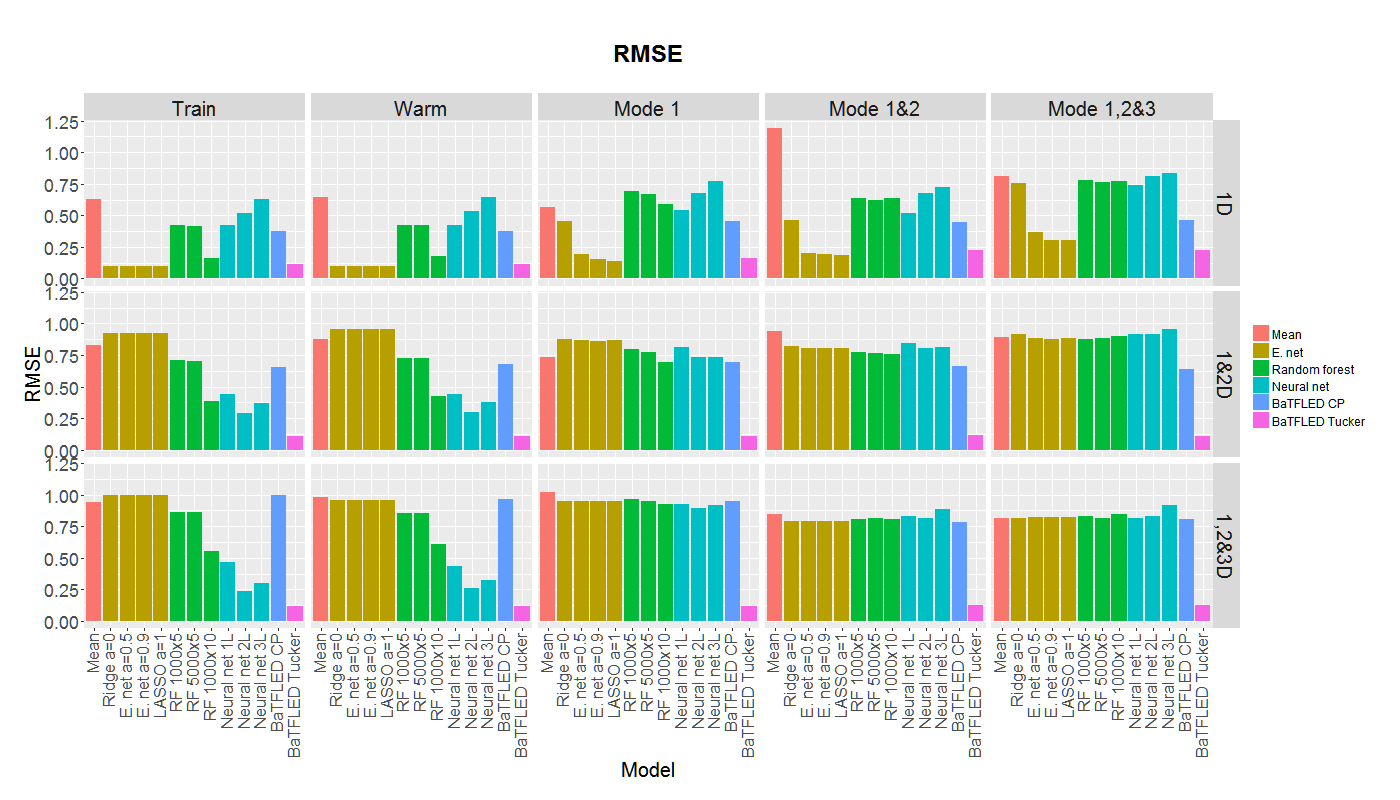}
  \caption{mean normalized RMSE performance on simulated data over 10 replicates.}
\end{figure}

\begin{figure}[!htbp]
  \centering
    \includegraphics[width=\textwidth]{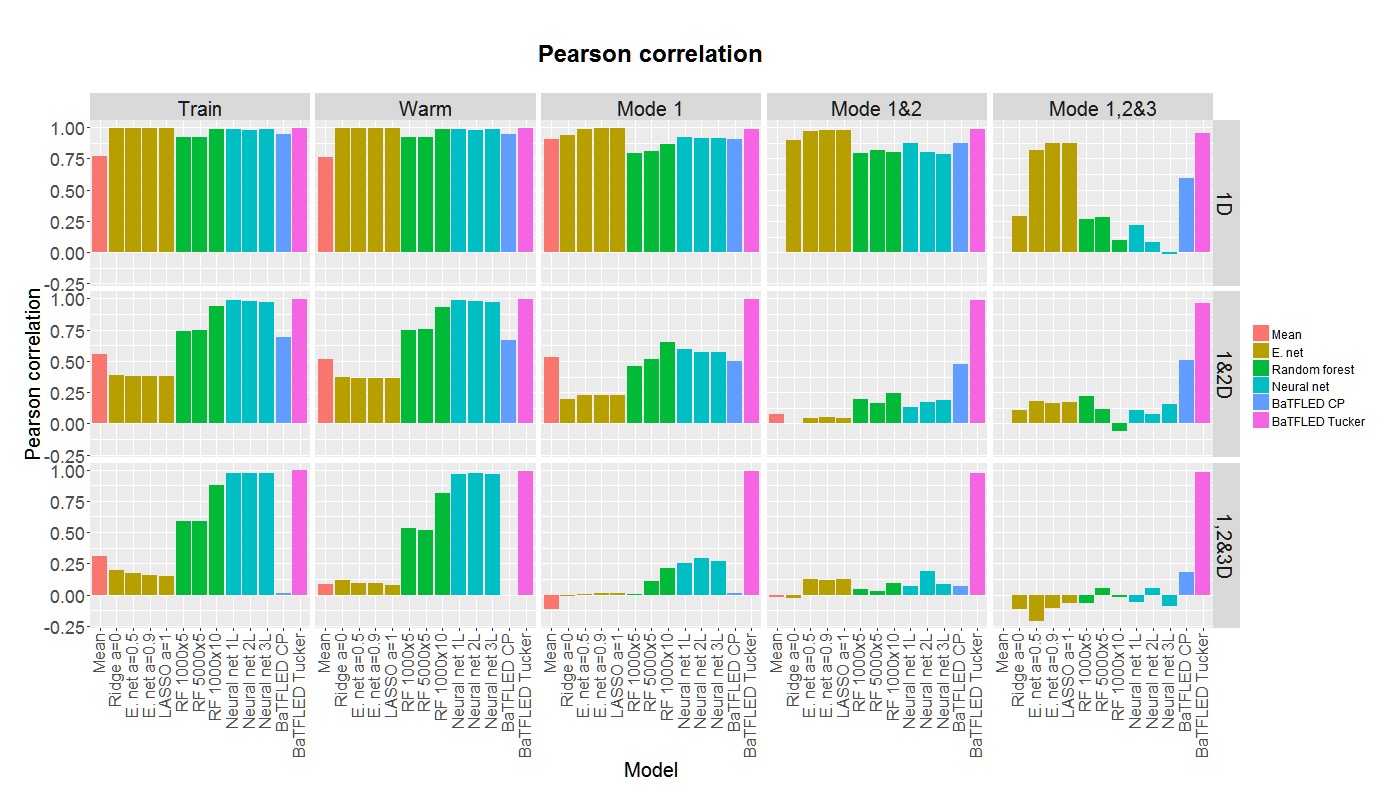}
  \caption{Mean Pearson correlation performance on simulated data over 10 replicates.}
\end{figure}

\begin{figure}[!htbp]
  \centering
    \includegraphics[width=\textwidth]{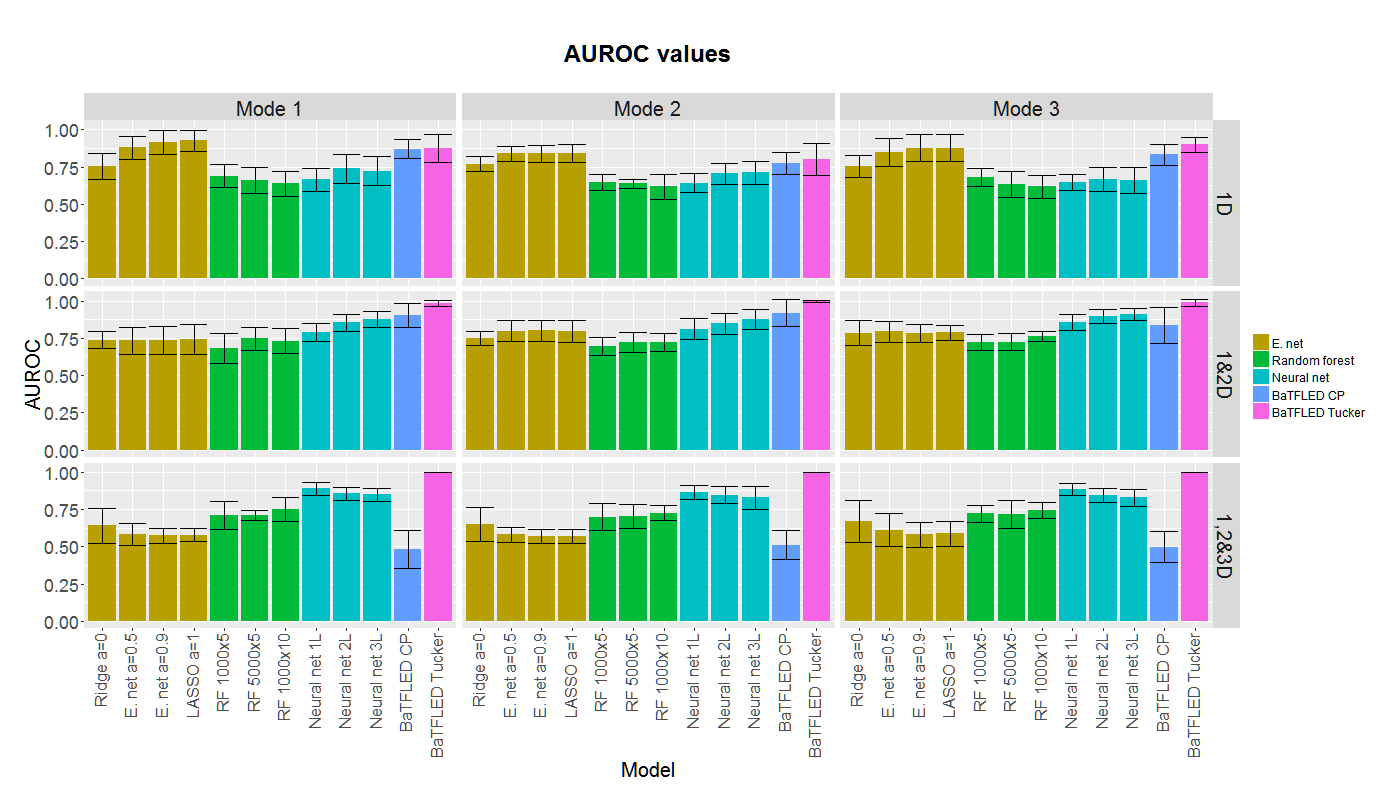}
  \caption{Area under the receiver operator characteristic curve (AUROC) for simulated data (mean over 10 replicates with error bars showing sd).}
\end{figure}

\subsection{DREAM data}

When running on DREAM data separate predictors are trained for each dose for elastic net, random forest models and neural net models. The inputs for each method are vectors containing the $733$ features for cell lines and $433$ features for drugs concatenated together. Elastic net models were run using the R package `glmnet' and while the `h2o' package was used for random forest and neural net training. The same model parameters that were used on the simulated data were tested here.

For cross-validation runs, BaTFLED models were run for 200 iterations with strong sparsity priors ($\alpha=10^{-10}$ and $\beta=10^{10}$) on the first round. The second round was run using the union of the top 15\% of predictors across folds (278 cell line features and 163 drug features for CP models, 152 cell line features and 105 drug features for Tucker models), for 40 iterations, without encouraging sparsity ($\alpha=1$ and $\beta=1$). Tucker models use a $10$x$10$x$10$ core and CP models have a latent dimension of $200$. 

For the final testing on the 17 held-out cell lines, BaTFLED models were run for 120 iterations for both rounds with sparsity encouraged only on first round. Again the union of the top 15\% of predictors across replicates were used for the second round (191 cell line features and 102 drug features for CP models and 330 cell line features and 172 drug features for Tucker models). Elastic net, random forest and neural net models were run as above.

All models were run on Intel Xeon processors in parallel when possible. The elastic net models run on a single core take a few minutes to complete while random forest and neural net models take 10-20 minutes on the simulated data while using 24 cores. On the DREAM data, run time was similar for the random forest models, but significantly longer for neural net models with some taking up to 14 hours. The BaTFLED models tested here run in 10-20 minutes for simulated data and up to 8 hours on the DREAM data using 16 cores. However, the code has not been optimized very throughly yet, so significant improvements are likely possible.

\begin{figure}[!htbp]
  \centering
    \includegraphics[width=\textwidth]{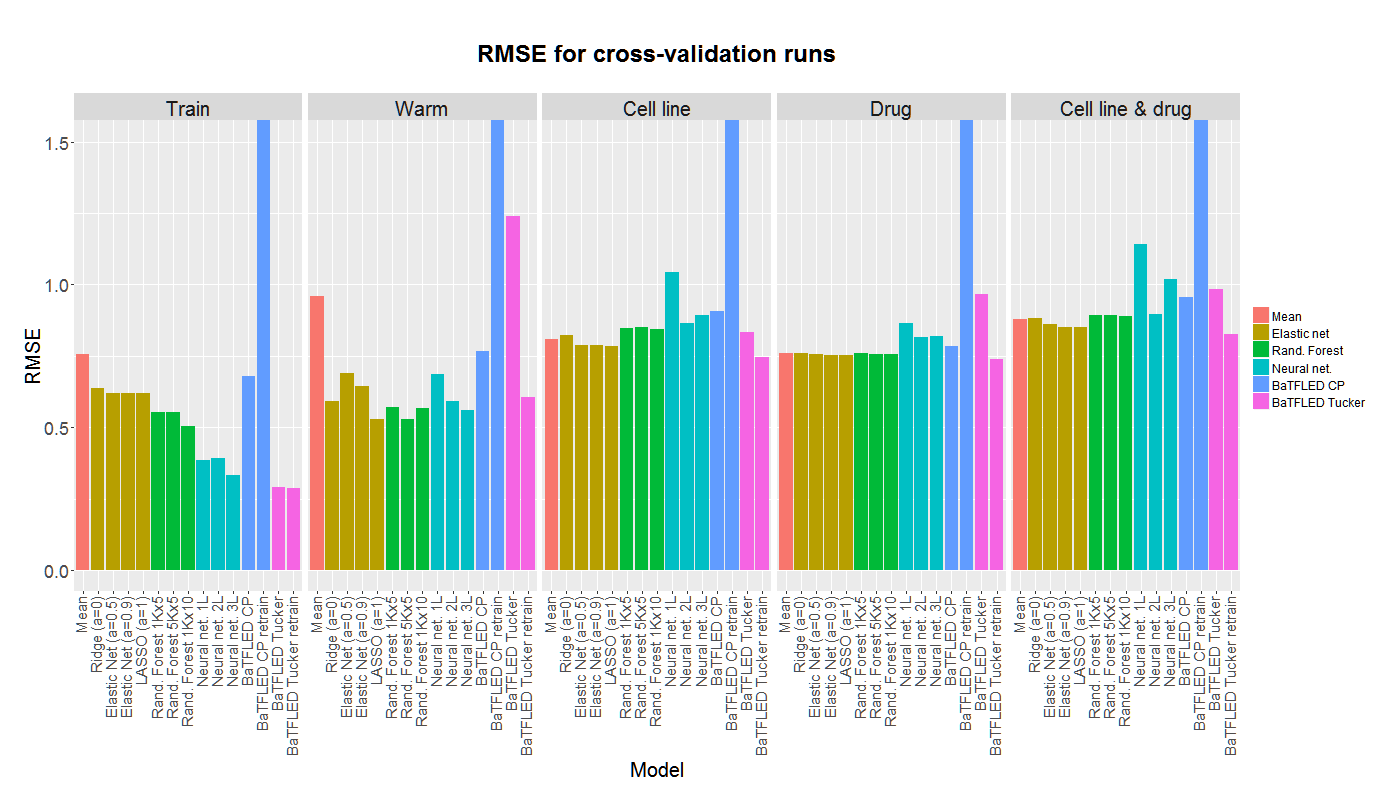}
  \caption{Mean normalized RMSE performance for cross-validation runs on DREAM data.}
\end{figure}

\begin{figure}[!htbp]
  \centering
    \includegraphics[width=\textwidth]{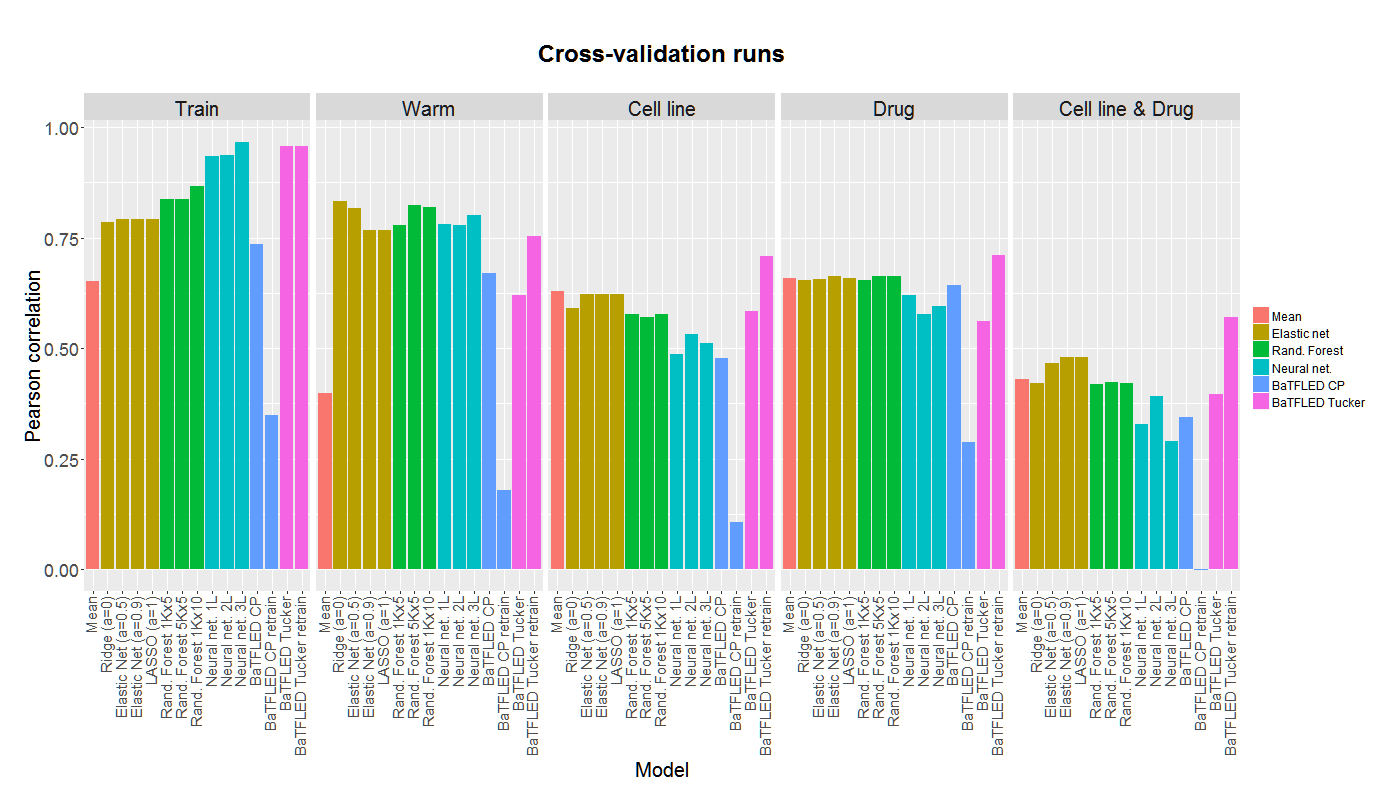}
  \caption{Mean Pearson correlation performance for cross-validation runs on DREAM data.}
\end{figure}

\begin{figure}[!htbp]
  \centering
    \includegraphics[width=\textwidth]{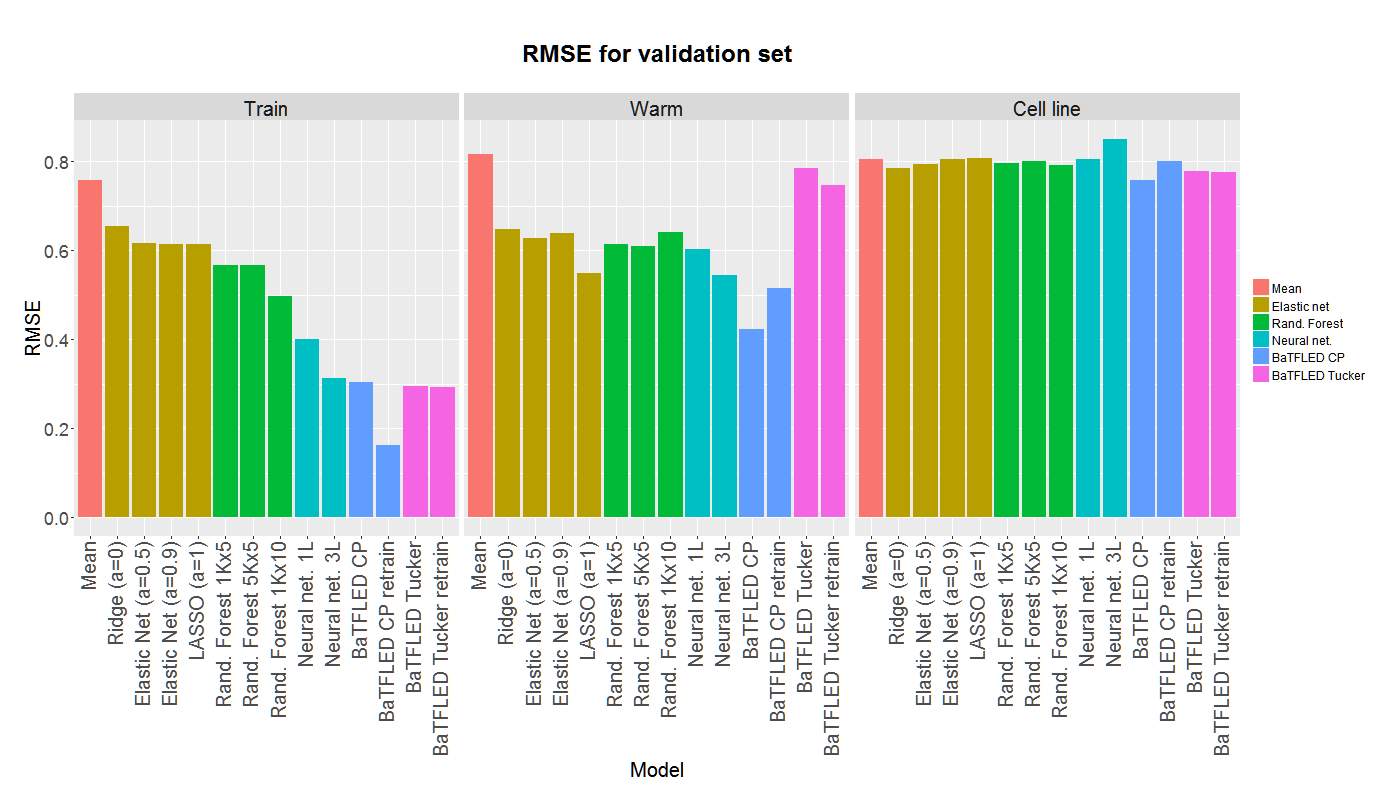}
  \caption{Mean normalized RMSE performance for DREAM data on final validation set over ten replicates.}
\end{figure}

\begin{figure}[!htbp]
  \centering
    \includegraphics[width=\textwidth]{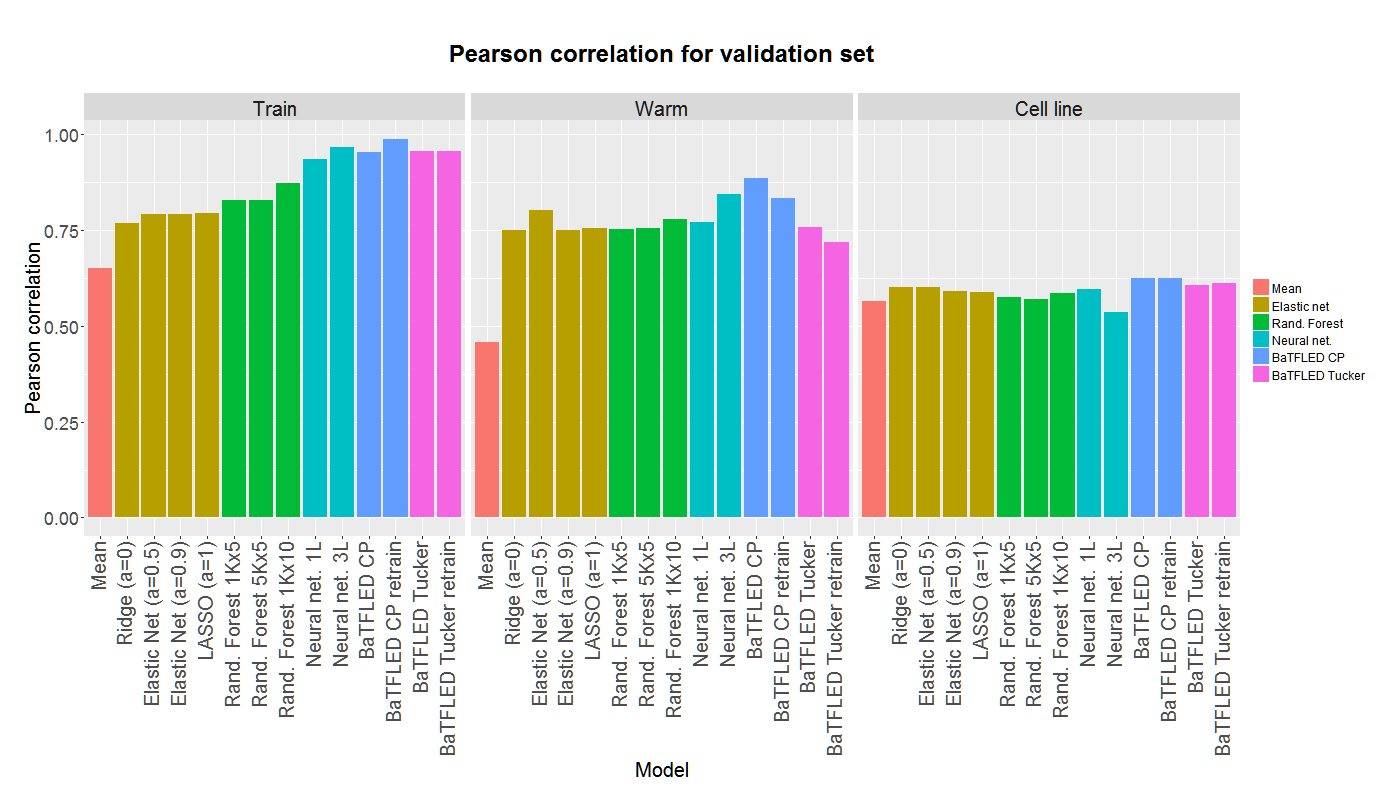}
  \caption{Mean Pearson correlation performance for DREAM data on final validation set over ten replicates.}
\end{figure}

\begin{table}[!htbp] 
  \caption{Mean normalized RMSE performance on simulated three-mode datasets relative to training set standard deviation across ten replicates (sd)}
  \label{simulated}
  \centering
  \resizebox*{\textwidth}{!}{
  \begin{tabular}{lccccccccc}
  \toprule
    & & & \multicolumn{7}{c}{Cold start prediction:}                   \\
    \cmidrule{4-10} 
        & Train & Warm &  Mode 1 & Mode 2 & Mode 3 & Mode 1\&2 & Mode 1\&3 & Mode 2\&3 & Mode 1,2\&3 \\
    \cmidrule{2-10}
    & \multicolumn{9}{c}{Data generated with only 1D interactions} \\
    \cmidrule{2-10} 
    Mean                       & 0.56(0.14) & 0.59(0.15) & 0.57(0.44) & 0.36(0.15) & 0.52(0.28) & 1.19(0.50) & 1.29(0.56) & 1.16(0.30) & 0.81(0.20) \\
    LASSO $\alpha=1$  & \textbf{0.10}(0.00) & \textbf{0.10}(0.01) & \textbf{0.14}(0.04) & \textbf{0.18}(0.10) & 0.27(0.19) & \textbf{0.18}(0.07) & 0.31(0.18) & 0.30(0.16) & 0.30(0.16) \\ 
    E. Net $\alpha=0.9$ & \textbf{0.10}(0.00) & \textbf{0.10}(0.01) & 0.15(0.04) & \textbf{0.18}(0.10) & 0.28(0.21) & 0.19(0.06) & 0.32(0.19) & 0.31(0.17) & 0.31(0.17) \\ 
    E. Net $\alpha=0.5$ & \textbf{0.10}(0.00) & \textbf{0.10}(0.01) & 0.19(0.04) & 0.20(0.11) & 0.32(0.28) & 0.20(0.06) & 0.39(0.25) & 0.37(0.25) & 0.37(0.25) \\ 
    Ridge $\alpha=0$  & \textbf{0.10}(0.00) & \textbf{0.10}(0.01) & 0.45(0.31) & 0.29(0.12) & 0.55(0.28) & 0.46(0.18) & 0.75(0.29) & 0.65(0.30) & 0.76(0.23) \\ 
    R. Forest 1Kx5 & 0.42(0.04) & 0.42(0.04) & 0.70(0.34) & 0.47(0.13) & 0.73(0.29) & 0.63(0.22) & 0.83(0.20) & 0.65(0.20) & 0.78(0.14) \\ 
    R. Forest 5Kx5 & 0.42(0.05) & 0.42(0.05) & 0.67(0.30) & 0.48(0.14) & 0.70(0.27) & 0.62(0.17) & 0.79(0.20) & 0.66(0.25) & 0.77(0.20) \\ 
    R. Forest 1Kx10 & 0.16(0.01) & 0.18(0.01) & 0.59(0.35) & 0.40(0.16) & 0.58(0.30) & 0.63(0.22) & 0.76(0.22) & 0.62(0.21) & 0.77(0.18) \\ 
    Neural net 1L.    & 0.43(0.05) & 0.42(0.06) & 0.54(0.31) & 0.50(0.08) & 0.58(0.25) & 0.52(0.20) & 0.74(0.24) & 0.71(0.27) & 0.74(0.19) \\ 
    Neural net 2L.    & 0.52(0.02) & 0.53(0.03) & 0.67(0.31) & 0.62(0.10) & 0.59(0.24) & 0.68(0.18) & 0.78(0.25) & 0.77(0.28) & 0.81(0.22) \\ 
    Neural net 3L.    & 0.63(0.03) & 0.65(0.04) & 0.77(0.28) & 0.68(0.07) & 0.72(0.21) & 0.73(0.16) & 0.85(0.22) & 0.82(0.23) & 0.84(0.18) \\ 
    BaTFLED CP.          & 0.37(0.04) & 0.38(0.04) & 0.46(0.15) & 0.42(0.11) & 0.46(0.13) & 0.45(0.20) & 0.52(0.15) & 0.45(0.11) & 0.46(0.18) \\ 
    BaTFLED Tucker   & 0.11(0.00) & 0.11(0.01) & 0.16(0.05) & 0.20(0.16) & \textbf{0.15}(0.03) & 0.22(0.13) & \textbf{0.18}(0.05) & \textbf{0.22}(0.14) & \textbf{0.23}(0.12) \\ 
    \cmidrule{2-10}
    & \multicolumn{9}{c}{Data generated with 1D \& 2D interactions} \\
    \cmidrule{2-10} 
    Mean                    & 0.83(0.05) & 0.88(0.07) & 0.74(0.26) & 0.78(0.22) & 0.84(0.39) & 0.94(0.30) & 1.09(0.31) & 1.37(0.34) & 0.89(0.39) \\ 
    LASSO $\alpha=1$  & 0.92(0.04) & 0.95(0.08) & 0.86(0.19) & 0.93(0.15) & 1.05(0.27) & 0.81(0.23) & 0.86(0.24) & 1.13(0.36) & 0.88(0.37) \\ 
    E. Net $\alpha=0.9$  & 0.92(0.04) & 0.96(0.08) & 0.86(0.19) & 0.93(0.15) & 1.05(0.27) & 0.80(0.23) & 0.86(0.23) & 1.13(0.36) & 0.88(0.36) \\ 
    E. Net $\alpha=0.5$  & 0.92(0.04) & 0.95(0.08) & 0.87(0.19) & 0.93(0.15) & 1.05(0.28) & 0.81(0.23) & 0.86(0.23) & 1.14(0.36) & 0.89(0.36) \\ 
    Ridge $\alpha=0$   & 0.92(0.04) & 0.95(0.08) & 0.87(0.19) & 0.95(0.15) & 1.06(0.28) & 0.82(0.26) & 0.85(0.20) & 1.18(0.38) & 0.92(0.40) \\ 
    R. Forest 1Kx5 &  0.71(0.08) & 0.73(0.09) & 0.79(0.18) & 0.84(0.17) & 1.01(0.35) & 0.77(0.22) & 0.86(0.21) & 1.11(0.33) & 0.88(0.34) \\ 
    R. Forest 5Kx5 & 0.71(0.08) & 0.72(0.09) & 0.78(0.18) & 0.83(0.17) & 1.01(0.36) & 0.77(0.23) & 0.83(0.22) & 1.08(0.30) & 0.88(0.33) \\ 
    R. Forest 1Kx10 & 0.38(0.04) & 0.42(0.05) & 0.69(0.20) & 0.73(0.18) & 0.93(0.41) & 0.75(0.21) & 0.82(0.22) & 1.05(0.31) & 0.90(0.35) \\ 
    Neural net 1L.    & 0.44(0.03) & 0.44(0.04) & 0.81(0.25) & 0.77(0.23) & 0.97(0.35) & 0.84(0.28) & 0.90(0.21) & 1.09(0.35) & 0.91(0.42) \\ 
    Neural net 2L.    & 0.29(0.02) & 0.30(0.02) & 0.74(0.27) & 0.71(0.22) & 0.87(0.38) & 0.80(0.30) & 0.86(0.22) & 1.09(0.33) & 0.91(0.45) \\ 
    Neural net 3L.    & 0.37(0.04) & 0.37(0.04) & 0.73(0.27) & 0.72(0.22) & 0.84(0.35) & 0.81(0.32) & 0.86(0.23) & 1.05(0.31) & 0.96(0.50) \\ 
   BaTFLED CP.         & 0.66(0.18) & 0.68(0.18) & 0.70(0.17) & 0.67(0.19) & 0.77(0.29) & 0.66(0.24) & 0.71(0.15) & 0.81(0.36) & 0.64(0.27) \\ 
    BaTFLED Tucker   & \textbf{0.11}(0.01) & \textbf{0.11}(0.01) & \textbf{0.11}(0.01) & \textbf{0.11}(0.01) & \textbf{0.11}(0.01) & \textbf{0.11}(0.01) & \textbf{0.12}(0.02) & \textbf{0.12}(0.02) & \textbf{0.11}(0.04) \\ 
    \cmidrule{2-10}
    & \multicolumn{9}{c}{Data generated with 1D, 2D \& 3D interactions} \\
    \cmidrule{2-10} 
    Mean                      & 0.94(0.04) & 0.98(0.05) & 1.02(0.30) & 0.80(0.37) & 1.05(0.32) & 0.85(0.43) & 1.13(0.47) & 0.89(0.49) & 0.81(0.58) \\ 
    LASSO $\alpha=1$  & 0.99(0.01) & 0.96(0.04) & 0.95(0.29) & 0.82(0.36) & 1.08(0.35) & 0.79(0.44) & 1.06(0.49) & 0.83(0.51) & 0.82(0.58) \\ 
    E. Net $\alpha=0.9$ & 1.00(0.01) & 0.96(0.04) & 0.95(0.29) & 0.82(0.36) & 1.08(0.35) & 0.79(0.44) & 1.06(0.49) & 0.83(0.51) & 0.82(0.58) \\ 
    E. Net $\alpha=0.5$ & 0.99(0.01) & 0.96(0.04) & 0.95(0.29) & 0.82(0.36) & 1.08(0.35) & 0.79(0.43) & 1.06(0.49) & 0.83(0.51) & 0.82(0.58) \\ 
    Ridge $\alpha=0$    & 1.00(0.01) & 0.96(0.04) & 0.95(0.29) & 0.82(0.36) & 1.08(0.35) & 0.79(0.44) & 1.06(0.49) & 0.83(0.51) & 0.81(0.58) \\ 
    R. Forest 1Kx5 & 0.86(0.03) & 0.85(0.04) & 0.96(0.30) & 0.78(0.34) & 1.07(0.34) & 0.81(0.40) & 1.08(0.48) & 0.82(0.49) & 0.83(0.55) \\ 
    R. Forest 5Kx5 & 0.86(0.02) & 0.86(0.04) & 0.95(0.27) & 0.80(0.34) & 1.07(0.33) & 0.81(0.39) & 1.06(0.47) & 0.83(0.49) & 0.81(0.56) \\ 
    R. Forest 1Kx10 & 0.55(0.05) & 0.61(0.07) & 0.92(0.25) & 0.78(0.32) & 1.05(0.31) & 0.81(0.39) & 1.06(0.44) & 0.84(0.46) & 0.85(0.55) \\ 
    Neural net 1L.    & 0.46(0.04) & 0.43(0.03) & 0.92(0.27) & 0.92(0.25) & 1.07(0.28) & 0.83(0.42) & 1.03(0.45) & 0.87(0.48) & 0.81(0.56) \\ 
    Neural net 2L.    & 0.24(0.01) & 0.26(0.02) & 0.89(0.25) & 0.81(0.32) & 1.10(0.26) & 0.81(0.43) & 1.04(0.45) & 0.86(0.47) & 0.83(0.56) \\ 
    Neural net 3L.    & 0.30(0.04) & 0.33(0.03) & 0.92(0.28) & 0.82(0.32) & 1.10(0.28) & 0.89(0.44) & 1.03(0.43) & 0.88(0.45) & 0.92(0.56) \\ 
    BaTFLED CP.            & 1.00(0.00) & 0.96(0.04) & 0.95(0.29) & 0.83(0.35) & 1.08(0.35) & 0.78(0.45) & 1.05(0.50) & 0.83(0.51) & 0.81(0.59) \\ 
    BaTFLED Tucker & \textbf{0.12}(0.03) & \textbf{0.12}(0.02) & \textbf{0.12}(0.03) & \textbf{0.13}(0.04) & \textbf{0.12}(0.02) & \textbf{0.13}(0.04) & \textbf{0.12}(0.02) & \textbf{0.13}(0.04) & \textbf{0.12}(0.04) \\ 
\bottomrule
  \end{tabular}}
\end{table}

\begin{table}[!htbp]
  \caption{Mean Pearson correlation performance on simulated three-mode datasets across ten replicates (sd)}
  \label{simulated}
  \centering
  \resizebox*{\textwidth}{!}{
  \begin{tabular}{lccccccccc}
  \toprule
    & & & \multicolumn{7}{c}{Cold start prediction:}                   \\
    \cmidrule{4-10} 
        & Train & Warm &  Mode 1 & Mode 2 & Mode 3 & Mode 1\&2 & Mode 1\&3 & Mode 2\&3 & Mode 1,2\&3 \\
    \cmidrule{2-10}
    & \multicolumn{9}{c}{Data generated with only 1D interactions} \\
    \cmidrule{2-10} 
    Mean                       & 0.77(0.08) & 0.76(0.08) & 0.90(0.11) & 0.97(0.03) & 0.85(0.15) & 0.00(0.42) & 0.01(0.51) & -0.12(0.39) & NA \\ 
    LASSO $\alpha=1$  & \textbf{1.00}(0.00) & \textbf{1.00}(0.00) & \textbf{0.99}(0.00) & \textbf{0.99}(0.01) & 0.94(0.13) & 0.97(0.03) & 0.90(0.20) & 0.92(0.13) & 0.88(0.20) \\ 
    E. Net $\alpha=0.9$ & \textbf{1.00}(0.00) & \textbf{1.00}(0.00) & \textbf{0.99}(0.01) & \textbf{0.99}(0.01) & 0.93(0.13) & 0.97(0.03) & 0.90(0.20) & 0.92(0.13) & 0.87(0.21) \\ 
    E. Net $\alpha=0.5$ & \textbf{1.00}(0.00) & \textbf{1.00}(0.00) & \textbf{0.99}(0.01) & \textbf{0.99}(0.01) & 0.91(0.16) & 0.97(0.04) & 0.86(0.22) & 0.89(0.16) & 0.82(0.23) \\ 
    Ridge $\alpha=0$  & \textbf{1.00}(0.00) & \textbf{1.00}(0.00) & 0.94(0.08) & 0.97(0.02) & 0.83(0.19) & 0.89(0.08) & 0.64(0.26) & 0.75(0.26) & 0.29(0.55) \\ 
    R. Forest 1Kx5 & 0.92(0.02) & 0.92(0.02) & 0.79(0.16) & 0.93(0.04) & 0.75(0.16) & 0.79(0.15) & 0.39(0.56) & 0.74(0.32) & 0.26(0.61) \\ 
    R. Forest 5Kx5 & 0.92(0.01) & 0.92(0.02) & 0.81(0.11) & 0.93(0.04) & 0.75(0.16) & 0.81(0.10) & 0.42(0.59) & 0.75(0.33) & 0.28(0.67) \\ 
    R. Forest 1Kx10 & 0.99(0.00) & 0.99(0.00) & 0.86(0.15) & 0.96(0.03) & 0.84(0.15) & 0.80(0.13) & 0.48(0.46) & 0.76(0.26) & 0.10(0.63) \\ 
    Neural net 1L.    & 0.99(0.00) & 0.99(0.00) & 0.92(0.08) & 0.96(0.03) & 0.85(0.15) & 0.87(0.08) & 0.64(0.26) & 0.78(0.19) & 0.21(0.56) \\ 
    Neural net 2L.    & 0.98(0.00) & 0.98(0.00) & 0.91(0.15) & 0.94(0.05) & 0.84(0.15) & 0.80(0.21) & 0.56(0.44) & 0.69(0.33) & 0.08(0.63) \\ 
    Neural net 3L.    & 0.99(0.00) & 0.99(0.00) & 0.91(0.13) & 0.95(0.04) & 0.84(0.15) & 0.78(0.23) & 0.38(0.50) & 0.64(0.35) & -0.02(0.69) \\ 
    BaTFLED CP.          & 0.95(0.01) & 0.95(0.01) & 0.90(0.09) & 0.94(0.03) & 0.88(0.09) & 0.87(0.13) & 0.76(0.26) & 0.86(0.11) & 0.59(0.55) \\ 
    BaTFLED Tucker     & 0.99(0.00) & 0.99(0.00) & \textbf{0.99}(0.01) & \textbf{0.99}(0.00) & \textbf{0.99}(0.01) & \textbf{0.98}(0.02) & \textbf{0.97}(0.02) & \textbf{0.98}(0.01) & \textbf{0.95}(0.06) \\ 
    \cmidrule{2-10}
    & \multicolumn{9}{c}{Data generated with 1D \& 2D interactions} \\
    \cmidrule{2-10} 
    Mean                    & 0.55(0.08) & 0.51(0.09) & 0.53(0.29) & 0.62(0.14) & 0.63(0.23) & 0.08(0.34) & 0.00(0.23) & -0.08(0.15) & NA \\ 
    LASSO $\alpha=1$  & 0.38(0.09) & 0.36(0.13) & 0.23(0.29) & 0.26(0.35) & 0.26(0.25) & 0.04(0.31) & 0.17(0.44) & 0.25(0.35) & 0.17(0.56) \\ 
    E. Net $\alpha=0.9$  & 0.38(0.09) & 0.36(0.13) & 0.23(0.29) & 0.26(0.35) & 0.26(0.24) & 0.05(0.30) & 0.17(0.44) & 0.26(0.35) & 0.17(0.57) \\ 
    E. Net $\alpha=0.5$  & 0.38(0.09) & 0.36(0.13) & 0.22(0.28) & 0.26(0.35) & 0.26(0.25) & 0.04(0.30) & 0.17(0.44) & 0.25(0.35) & 0.18(0.56) \\ 
    Ridge $\alpha=0$   & 0.39(0.09) & 0.37(0.13) & 0.20(0.30) & 0.26(0.34) & 0.25(0.26) & 0.01(0.30) & 0.23(0.39) & 0.21(0.39) & 0.11(0.32) \\ 
    R. Forest 1Kx5 & 0.74(0.06) & 0.75(0.08) & 0.46(0.20) & 0.46(0.34) & 0.40(0.28) & 0.19(0.36) & 0.24(0.36) & 0.33(0.37) & 0.22(0.46) \\ 
    R. Forest 5Kx5 & 0.75(0.06) & 0.75(0.07) & 0.51(0.20) & 0.44(0.41) & 0.38(0.35) & 0.16(0.39) & 0.35(0.30) & 0.34(0.43) & 0.12(0.51) \\ 
    R. Forest 1Kx10 & 0.94(0.01) & 0.93(0.02) & 0.65(0.22) & 0.64(0.29) & 0.53(0.29) & 0.24(0.53) & 0.34(0.41) & 0.47(0.45) & -0.06(0.66) \\ 
    Neural net 1L.    & 0.98(0.00) & 0.98(0.00) & 0.60(0.19) & 0.69(0.16) & 0.62(0.22) & 0.13(0.38) & 0.28(0.21) & 0.40(0.40) & 0.11(0.45) \\ 
    Neural net 2L.    & 0.98(0.00) & 0.98(0.00) & 0.57(0.18) & 0.71(0.16) & 0.63(0.20) & 0.17(0.34) & 0.22(0.27) & 0.51(0.23) & 0.07(0.36) \\ 
    Neural net 3L.    & 0.97(0.00) & 0.97(0.01) & 0.57(0.19) & 0.70(0.17) & 0.64(0.20) & 0.19(0.40) & 0.24(0.25) & 0.54(0.27) & 0.16(0.43) \\ 
   BaTFLED CP.         & 0.69(0.26) & 0.67(0.31) & 0.50(0.34) & 0.67(0.28) & 0.66(0.27) & 0.47(0.35) & 0.48(0.28) & 0.70(0.26) & 0.51(0.50) \\ 
    BaTFLED Tucker   & \textbf{0.99}(0.00) & \textbf{0.99}(0.00) & \textbf{0.99}(0.00) & \textbf{0.99}(0.00) & \textbf{0.99}(0.00) & \textbf{0.99}(0.01) & \textbf{0.99}(0.01) & \textbf{0.99}(0.01) & \textbf{0.96}(0.08) \\ 
    \cmidrule{2-10}
    & \multicolumn{9}{c}{Data generated with 1D, 2D \& 3D interactions} \\
    \cmidrule{2-10} 
    Mean                       & 0.31(0.11)  & 0.08(0.12) & -0.11(0.50) & 0.22(0.40) & 0.23(0.25) & -0.02(0.16) & -0.05(0.14) & 0.03(0.08) &  NA \\ 
    LASSO $\alpha=1$*   & 0.15(0.05) & 0.08(0.08) & 0.02(0.17) & 0.20(0.12) & 0.01(0.17) & 0.12(0.17) & -0.19(0.23) & 0.16(0.25) & -0.07(0.35)\\ 
    E. Net $\alpha=0.9$* & 0.16(0.06) & 0.09(0.09) & 0.01(0.19) & 0.17(0.12) & 0.00(0.19) & 0.12(0.19) & -0.23(0.25) & 0.12(0.26) & -0.11(0.37) \\ 
    E. Net $\alpha=0.5$* & 0.17(0.05) & 0.10(0.09) & 0.00(0.20) & 0.18(0.12) & 0.01(0.20) & 0.13(0.20) & -0.29(0.25) & 0.18(0.24) & -0.21(0.34) \\ 
    Ridge $\alpha=0$*  & 0.20(0.05) & 0.11(0.10) & -0.01(0.22) & 0.11(0.18) & 0.01(0.25) & -0.02(0.30) & -0.20(0.35) & 0.03(0.36) & -0.11(0.31) \\ 
    R. Forest 1Kx5 & 0.59(0.06) & 0.53(0.11) & 0.01(0.30) & 0.27(0.31) & 0.13(0.20) & 0.05(0.25) & -0.04(0.26) & 0.18(0.34) & -0.07(0.43) \\ 
    R. Forest 5Kx5 & 0.59(0.06) & 0.51(0.11) & 0.11(0.35) & 0.23(0.33) & 0.14(0.17) & 0.03(0.32) & 0.04(0.26) & 0.16(0.30) & 0.05(0.43) \\ 
    R. Forest 1Kx10 & 0.88(0.03) & 0.81(0.06) & 0.22(0.49) & 0.33(0.34) & 0.24(0.24) & 0.09(0.28) & 0.04(0.37) & 0.17(0.32) & -0.02(0.46) \\ 
    Neural net 1L.    & 0.97(0.00) & 0.96(0.01) & 0.25(0.50) & 0.39(0.39) & 0.21(0.27) & 0.07(0.35) & 0.12(0.37) & 0.05(0.26) & -0.06(0.26) \\ 
    Neural net 2L.    & 0.97(0.00) & 0.97(0.00) & 0.30(0.45) & 0.38(0.38) & 0.22(0.19) & 0.19(0.36) & 0.14(0.31) & 0.09(0.33) & 0.06(0.30) \\ 
    Neural net 3L.    & 0.97(0.00) & 0.96(0.00) & 0.27(0.47) & 0.35(0.44) & 0.21(0.24) & 0.09(0.36) & 0.19(0.37) & 0.12(0.34) & -0.09(0.37) \\ 
    BaTFLED CP.            & 0.01(0.03) & 0.00(0.07) & 0.01(0.06) & -0.02(0.05) & 0.02(0.07) & 0.07(0.20) & 0.02(0.19) & 0.00(0.16) & 0.18(0.36) \\ 
    BaTFLED Tucker  & \textbf{0.99}(0.00) & \textbf{0.99}(0.00) & \textbf{0.99}(0.01) & \textbf{0.98}(0.01) & \textbf{0.99}(0.00) & \textbf{0.97}(0.04) & \textbf{0.99}(0.02) & \textbf{0.98}(0.01) & \textbf{0.98}(0.03) \\ 
\bottomrule
    \multicolumn{9}{l}{* Elastic net set all weights to zero in 4-5 of 10 runs.} \\
  \end{tabular}}
\end{table}

\begin{table}[h]
  \caption{Mean area under the receiver operator characteristic curve (AUROC) for selecting predictors in simulated three-mode data over 10 replicates (sd)}
  \label{simulated_preds}
  \centering
  \resizebox*{\textwidth}{!}{
  \begin{tabular}{lcccccccccccc}
    \toprule
    & \multicolumn{1}{p{1cm}}{\centering LASSO $\alpha=1$}
    & \multicolumn{1}{p{1cm}}{\centering E. Net $\alpha=.9$}
    & \multicolumn{1}{p{1cm}}{\centering E. Net $\alpha=.5$} 
    & \multicolumn{1}{p{1cm}}{\centering Ridge $\alpha=0$} 
    & \multicolumn{1}{p{1cm}}{\centering RF 1Kx5} 
    & \multicolumn{1}{p{1cm}}{\centering RF 5Kx5} 
    & \multicolumn{1}{p{1cm}}{\centering RF 1Kx10} 
    & \multicolumn{1}{p{1cm}}{\centering Neural net 1L.} 
    & \multicolumn{1}{p{1cm}}{\centering Neural net 2L.} 
    & \multicolumn{1}{p{1cm}}{\centering Neural net 3L.} 
    & \multicolumn{1}{p{1.2cm}}{\centering BaTFLED CP.}
    & \multicolumn{1}{p{1.2cm}}{\centering BaTFLED Tucker} \\
    \cmidrule{2-13}
    & \multicolumn{12}{c}{Data generated with only 1D interactions} \\ 
    \cmidrule{2-13} 
    Mode 1  & \textbf{0.93}(0.07) & 0.92(0.08) & 0.88(0.08) & 0.75(0.09) & 0.69(0.08) & 0.66(0.09) & 0.64(0.08) & 0.66(0.07) & 0.74(0.10) & 0.72(0.10) & 0.87(0.06) & 0.87(0.09) \\ 
    Mode 2  & \textbf{0.84}(0.06) & \textbf{0.84}(0.06) & \textbf{0.84}(0.05) & 0.77(0.05) & 0.65(0.05) & 0.64(0.03) & 0.62(0.08) & 0.64(0.06) & 0.70(0.07) & 0.71(0.08) & 0.78(0.08) & 0.80(0.11) \\ 
    Mode 3  & 0.88(0.09) & 0.88(0.09) & 0.85(0.09) & 0.76(0.07) & 0.68(0.06) & 0.63(0.09) & 0.62(0.08) & 0.65(0.05) & 0.67(0.08) & 0.66(0.09) & 0.83(0.07) & \textbf{0.90}(0.05) \\ 
    & \multicolumn{12}{c}{Data generated with 1D \& 2D interactions} \\
    \cmidrule{2-13} 
    Mode 1  & 0.74(0.10) & 0.73(0.09) & 0.73(0.09) & 0.74(0.06) & 0.68(0.10) & 0.75(0.08) & 0.73(0.08) & 0.79(0.06) & 0.85(0.06) & 0.88(0.05) & 0.90(0.08) & \textbf{0.99}(0.02) \\
    Mode 2  & 0.80(0.08) & 0.80(0.07) & 0.80(0.07) & 0.75(0.05) & 0.70(0.06) & 0.72(0.07) & 0.72(0.06) & 0.81(0.07) & 0.85(0.07) & 0.88(0.07) & 0.92(0.09) & \textbf{1.00}(0.01) \\
    Mode 3  & 0.79(0.05) & 0.78(0.06) & 0.79(0.07) & 0.79(0.09) & 0.72(0.05) & 0.72(0.05) & 0.76(0.04) & 0.86(0.06) & 0.90(0.04) & 0.91(0.04) & 0.84(0.12) & \textbf{0.99}(0.02) \\
    & \multicolumn{12}{c}{Data generated with 1D, 2D \& 3D interactions} \\
    \cmidrule{2-13} 
    Mode 1  & 0.58(0.04) & 0.57(0.05) & 0.58(0.07) & 0.64(0.12) & 0.71(0.09) & 0.71(0.04) & 0.75(0.08) &  0.89(0.04) & 0.86(0.04) & 0.85(0.04) & 0.48(0.13) & \textbf{1.00}(0.00) \\
    Mode 2  & 0.57(0.05) & 0.57(0.05) & 0.58(0.05) & 0.65(0.12) & 0.70(0.09) & 0.70(0.08) & 0.73(0.05) & 0.87(0.05) & 0.85(0.06) & 0.83(0.08) & 0.51(0.10) & \textbf{1.00}(0.00) \\
    Mode 3  & 0.59(0.08) & 0.58(0.08) & 0.61(0.11) & 0.67(0.14) & 0.72(0.06) & 0.72(0.09) & 0.74(0.06) &  0.88(0.04) & 0.84(0.05) & 0.83(0.06) & 0.50(0.10) & \textbf{1.00}(0.00) \\
    \bottomrule
  \end{tabular}}
\end{table}

\begin{table}[!htbp]
  \caption{Mean normalized RMSE on DREAM dataset over ten replicates (sd).}
  \label{DREAM}
  \centering
  \resizebox*{\textwidth}{!}{
  \begin{tabular}{lccccc}
    \toprule
          & Training & Warm & Cell lines & Drugs 
          & \multicolumn{1}{p{1.5cm}}{\centering Cell lines \& drugs}  \\
    \cmidrule{2-6}
    & \multicolumn{5}{c}{First training round cross-validation} \\
    \cmidrule{2-6} 
    Mean                      & 0.76(0.01) & 0.96(0.18) & 0.81(0.23) & 0.76(0.07) & 0.88(0.24) \\ 
    LASSO $\alpha=1$   & 0.62(0.01) & \textbf{0.53}(0.10) & 0.79(0.19) & 0.75(0.07) & 0.85(0.19) \\ 
    E. Net $\alpha=0.9$ & 0.62(0.01) & 0.65(0.22) & 0.79(0.19) & 0.75(0.07) & 0.85(0.19) \\ 
    E. Net $\alpha=0.5$ & 0.62(0.01) & 0.69(0.17) & 0.79(0.20) & 0.76(0.06) & 0.86(0.19) \\ 
    Ridge $\alpha=0$     & 0.64(0.01) & 0.59(0.25) & 0.82(0.20) & 0.76(0.07) & 0.88(0.20) \\ 
    R. forest 1Kx5       & 0.55(0.01) & 0.57(0.21) & 0.85(0.21) & 0.76(0.07) & 0.89(0.21) \\ 
    R. forest 5Kx5       & 0.55(0.01) & \textbf{0.53}(0.20) & 0.85(0.21) & 0.76(0.07) & 0.89(0.21) \\ 
    R. forest 1Kx10     & 0.50(0.01) & 0.57(0.17) & 0.85(0.22) & 0.75(0.07) & 0.89(0.21) \\ 
    Neural net 1L.       & 0.38(0.03) & 0.69(0.18) & 1.04(0.20) & 0.86(0.12) & 1.14(0.27) \\ 
    Neural net 2L.       & 0.39(0.02) & 0.59(0.15) & 0.87(0.20) & 0.82(0.07) & 0.89(0.23) \\ 
    Neural net 3L.       & 0.33(0.02) & 0.56(0.21) & 0.89(0.20) & 0.82(0.05) & 1.02(0.19) \\ 
    BaTFLED CP.            & 0.68(0.03) & 0.77(0.30) & 0.91(0.18) & 0.78(0.06) & 0.96(0.21) \\ 
    BaTFLED Tucker       & \textbf{0.29}(0.02) & 1.24(0.79) & 0.83(0.19) & 0.97(0.12) & 0.99(0.25) \\ 
    & \multicolumn{5}{c}{Second training round cross-validation} \\
    \cmidrule{2-6} 
    BaTFLED CP             & 1.87(0.39) & 1.77(0.44) & 2.00(0.34) & 1.88(0.27) & 1.90(0.35) \\ 
    BaTFLED Tucker      & \textbf{0.29}(0.02) & 0.60(0.23) & \textbf{0.74}(0.12) & \textbf{0.74}(0.05) & \textbf{0.83}(0.26) \\ 
    & \multicolumn{5}{c}{First training round on 17 held out cell lines} \\
    \cmidrule{2-6} 
    Mean                        & 0.76(0.00) & 0.82(0.10) & 0.80(0.00) & NA & NA \\	
    LASSO $\alpha=1$   & 0.61(0.00) & 0.55(0.11) & 0.81(0.00) & NA & NA \\ 
    E. Net $\alpha=0.9$ & 0.61(0.00) & 0.64(0.17) & 0.81(0.00) & NA & NA \\ 
    E. Net $\alpha=0.5$  & 0.62(0.00) & 0.63(0.25) & 0.79(0.00) & NA & NA \\ 
    Ridge $\alpha=0$     & 0.66(0.00) & 0.65(0.17) & 0.78(0.00) &  NA & NA \\ 
    R. forest 1Kx5       & 0.57(0.00) & 0.61(0.17) & 0.80(0.01) & NA & NA  \\ 
    R. forest 5Kx5       & 0.57(0.00) & 0.61(0.17) & 0.80(0.01) & NA & NA \\ 
    R. forest 1Kx10     & 0.50(0.00) & 0.64(0.20) & 0.79(0.00) & NA & NA \\ 
    Neural net 1L.       & 0.40(0.02) & 0.60(0.17) & 0.81(0.01) & NA & NA \\  
    Neural net 2L.      & 0.37(0.02) & 0.61(0.10) & 0.82(0.00) & NA & NA \\  
    Neural net 3L.      & 0.31(0.01) & 0.55(0.14) & 0.85(0.01) & NA & NA \\  
    BaTFLED CP.            & 0.30(0.01) & \textbf{0.42}(0.14) & \textbf{0.76}(0.00) & NA & NA \\ 
    BaTFLED Tucker       & 0.29(0.00) & 0.79(0.65) & 0.78(0.01) & NA & NA \\
    & \multicolumn{5}{c}{Second training round on 17 held out cell lines} \\ 
    \cmidrule{2-6} 
    BaTFLED CP.            & \textbf{0.16}(0.00) & 0.52(0.10) & 0.80(0.01) & NA & NA \\ 
    BaTFLED Tucker       & 0.29(0.00) & 0.75(0.53) & 0.78(0.02) & NA & NA \\ 
    \bottomrule
  \end{tabular}}
\end{table}

\begin{table}[!htbp]
  \caption{Mean Pearson correlation on DREAM dataset over ten replicates (sd).}
  \label{DREAM}
  \centering
  \resizebox*{\textwidth}{!}{
  \begin{tabular}{lccccc}
    \toprule
           & Training & Warm & Cell lines & Drugs 
           & \multicolumn{1}{p{1.5cm}}{\centering Cell lines \& drugs}  \\
    \cmidrule{2-6}
    & \multicolumn{5}{c}{First training round cross-validation} \\
    \cmidrule{2-6} 
    Mean                      & 0.65(0.02) & 0.40(0.22) & 0.63(0.10) & 0.66(0.05) & 0.43(0.12) \\
    LASSO $\alpha=1$   & 0.79(0.01) & 0.77(0.12) & 0.62(0.09) & 0.66(0.09) & 0.48(0.14) \\ 
    E. Net $\alpha=0.9$ & 0.79(0.01) & 0.77(0.13) & 0.62(0.09) & 0.66(0.09) & 0.48(0.13) \\ 
    E. Net $\alpha=0.5$ & 0.79(0.01) & 0.82(0.11) & 0.62(0.09) & 0.66(0.09) & 0.47(0.14)  \\ 
    Ridge $\alpha=0$     & 0.78(0.01) & \textbf{0.83}(0.07) & 0.59(0.10) & 0.65(0.09) & 0.42(0.13) \\ 
    R. forest 1Kx5       & 0.84(0.01) & 0.78(0.11) & 0.58(0.13) & 0.65(0.09) & 0.42(0.17) \\ 
    R. forest 5Kx5       & 0.84(0.01) & 0.82(0.10) & 0.57(0.12) & 0.66(0.07) & 0.42(0.16) \\ 
    R. forest 1Kx10     & 0.87(0.01) & 0.82(0.06) & 0.58(0.13) & 0.66(0.08) & 0.42(0.16) \\ 
    Neural net 1L.       & 0.93(0.01) & 0.78(0.08) & 0.49(0.10) & 0.62(0.08) & 0.33(0.15) \\ 
    Neural net 2L.       & 0.94(0.01) & 0.78(0.05) & 0.53(0.10) & 0.58(0.08) & 0.39(0.20) \\ 
    Neural net 3L.       & \textbf{0.97}(0.01) & 0.80(0.12) & 0.51(0.13) & 0.60(0.08) & 0.29(0.20) \\ 
    BaTFLED CP.            & 0.73(0.03) & 0.67(0.11) & 0.48(0.09) & 0.64(0.06) & 0.34(0.13) \\ 
    BaTFLED Tucker       & 0.96(0.01) & 0.62(0.36) & 0.58(0.11) & 0.56(0.10) & 0.40(0.20) \\ 
    & \multicolumn{5}{c}{Second training round cross-validation} \\
    \cmidrule{2-6} 
    BaTFLED CP        & 0.35(0.15) & 0.18(0.35) & 0.11(0.25) & 0.29(0.18) & 0.00(0.35) \\ 
    BaTFLED Tucker  & 0.96(0.01) & 0.75(0.13) & \textbf{0.71}(0.07) & \textbf{0.71}(0.06) & \textbf{0.57}(0.30) \\ 
    & \multicolumn{5}{c}{First training round on 17 held out cell lines} \\
    \cmidrule{2-6} 
    Mean               & 0.65(0.00) & 0.46(0.20) & 0.57(0.00) & NA & NA \\	
    LASSO $\alpha=1$   & 0.79(0.00) & 0.75(0.14) & 0.59(0.00) & NA & NA \\ 
    E. Net $\alpha=0.9$  & 0.79(0.00) & 0.75(0.18) & 0.59(0.00) & NA & NA \\ 
    E. Net $\alpha=0.5$  & 0.79(0.00) & 0.80(0.14) & 0.60(0.00) & NA & NA \\ 
    Ridge $\alpha=0$     & 0.77(0.00) & 0.75(0.12) & 0.60(0.00) & NA & NA \\ 
    R. forest 1Kx5       & 0.83(0.00) & 0.75(0.18) & 0.57(0.01) & NA & NA \\ 
    R. forest 5Kx5       & 0.83(0.00) & 0.76(0.14) & 0.57(0.01) & NA & NA \\ 
    R. forest 1Kx10     & 0.87(0.00) & 0.78(0.12) & 0.58(0.00) & NA & NA \\ 
    Neural net 1L.      & 0.93(0.01) & 0.77(0.20) & 0.60(0.01) & NA & NA \\  
    Neural net 2L.      & 0.94(0.01) & 0.76(0.08) & 0.57(0.00) & NA & NA \\  
    Neural net 3L.      & 0.97(0.00) & 0.84(0.10) & 0.54(0.01) & NA & NA \\  
    BaTFLED CP.            & 0.95(0.00)  & \textbf{0.88}(0.08) & \textbf{0.62}(0.01) & NA & NA \\  
    BaTFLED Tucker       & 0.96(0.00) & 0.76(0.28) & 0.61(0.02) & NA & NA \\ 
    & \multicolumn{5}{c}{Second training round on 17 held out cell lines} \\
    \cmidrule{2-6} 
    BaTFLED CP.            & \textbf{0.99}(0.00) & 0.83(0.11) & \textbf{0.62}(0.00) & NA & NA \\ 
    BaTFLED Tucker       & 0.96(0.00) & 0.72(0.30) & 0.61(0.02) & NA & NA \\ 
    \bottomrule
  \end{tabular}}
\end{table}

\clearpage

\subsubsection*{Acknowledgments}

Thanks to Dr. Shannon McWeeney, Dr. Adam Margolin and Dr. Lucia Carbone.

\section*{References}

Aiello, S., Kraljevic, T., Maj, P. and with contributions from the H2O.ai team (2016). h2o: R Interface for H2O. R package version 3.10.0.8. https://CRAN.R-project.org/package=h2o

Friedman, J., Hastie, T.\ \& Tibshirani R.\ (2010). Regularization Paths for Generalized Linear Models via Coordinate Descent. {\it Journal of Statistical Software} {\bf 33}(1): 1-22. URL http://www.jstatsoft.org/v33/i01/

\end{document}